\documentclass[journal,10pt,twocolumn,twoside]{IEEEtran}

\usepackage{hyperref}
\usepackage{multirow}
\usepackage{multicol}
\usepackage{graphicx}
\usepackage{makecell}
\usepackage{color,soul}
\usepackage[normalem]{ulem}
\useunder{\uline}{\ul}{}
\usepackage{balance}

\begin{document}

\title{Deep Learning for Intelligent Demand Response and Smart Grids: A Comprehensive Survey}

\author{Prabadevi B, Quoc-Viet Pham, Madhusanka Liyanage, N Deepa, Mounik VVSS, Shivani Reddy, \\Praveen Kumar Reddy Maddikunta, Neelu Khare, Thippa Reddy Gadekallu, and Won-Joo Hwang

\thanks{Prabadevi B, N Deepa,  Praveen Kumar Reddy Maddikunta, Neelu Khare, Thippa Reddy Gadekallu are with the School of Information Technology, Vellore Institute of Technology, Tamil Nadu- 632014, India (email: \{prabadevi.b, deepa.rajesh, praveenkumarreddy, neelu.khare, thippareddy.g\}@vit.ac.in).}

\thanks{Quoc-Viet Pham is with the Research Institute of Computer, Information and Communication Pusan National University, Busan 46241, Korea (email: vietpq@pusan.ac.kr).}

\thanks{Madhusanka Liyanage is with the School of Computer Science, University Collage Dublin, Ireland and Centre for Wireless Communications, University of Oulu, Finland (email: madhusanka@ucd.ie, madhusanka.liyanage@oulu.fi).}

\thanks{Mounik VVSS is with the Computer Science and Engineering, Indian Institute of Technology Madras, India (email: cs17b031@smail.iitm.ac.in).}

\thanks{Shivani Reddy is with the School of  Electronics and Communication Engineering, Vellore Institute of Technology, Tamil Nadu- 632014, India (email: shivani.devulapally2017@vitstudent.ac.in).}

\thanks{Won-Joo Hwang is with the Department of Biomedical Convergence Engineering, Pusan National University, Yangsan 50612, Korea (e-mail: wjhwang@pusan.ac.kr).}

}


\IEEEtitleabstractindextext{
\begin{abstract}
Electricity is one of the mandatory commodities for mankind today. To address challenges and issues in the transmission of electricity through the traditional grid, the concepts of smart grids and demand response have been developed. In such systems, a large amount of data is generated daily from various sources such as power generation (e.g., wind turbines), transmission and distribution (microgrids and fault detectors), load management (smart meters and smart electric appliances). Thanks to recent advancements in big data and computing technologies, Deep Learning (DL) can be leveraged to learn the patterns from the generated data and predict the demand for electricity and peak hours. Motivated by the advantages of deep learning in smart grids, this paper sets to provide a comprehensive survey on the application of DL for intelligent smart grids and demand response. Firstly, we present the fundamental of DL, smart grids, demand response, and the motivation behind the use of DL. Secondly, we review the state-of-the-art applications of DL in smart grids and demand response, including electric load forecasting, state estimation, energy theft detection, energy sharing and trading. Furthermore, we illustrate the practicality of DL via various use cases and projects. Finally, we highlight the challenges presented in existing research works and highlight important issues and potential directions in the use of DL for smart grids and demand response. 
\end{abstract}

}

\maketitle
\IEEEdisplaynontitleabstractindextext
\IEEEpeerreviewmaketitle

\section{Introduction}
\label{Sec:Introduction}

\begin{table}[h!]
\centering
\caption{Summary of Important Acronym.}
\label{tab:Acronym}
\begin{tabular}{ll}
SG        & Smart Grid                                     \\
IoT       & Internet of Things                             \\
5G        & Fifth Generation                               \\
ML        & Machine Learning                               \\
DR        & Demand Response                                \\
DL        & Deep Learning                                  \\
RNN       & Recurrent Neural Networks                      \\
WPC       & Wireless Powered Communication                 \\
AI        & Artificial Intelligence                        \\
DNN       & Deep Neural Networks                           \\
ANN       & Artificial Neural Networks                     \\
CNN       & Convolutional Neural Network                   \\
LSTM      & Long Short-Term Memory                         \\
GRU       & Gated Recurrent Unit                           \\
IL        & Interruptible Load                             \\
DDQN      & Dueling Deep Q Network                         \\
STLF      & Short-Term Load Forecasting                    \\
EV        & Electric Vehicles                              \\
ARIMA     & Auto-Regressive Integrated Moving Average      \\
RU        & Rolling Update                                 \\
AM        & Attention Mechanism                            \\
Bi-LSTM   & Bidirectional-LSTM                             \\
MKL       & Multiple Kernel Learning                       \\
LM        & Levenberg-Marquardt                            \\
BR        & Bayesian Regularization                        \\
SVR       & Support Vector Regression                      \\
PV        & Photovoltaic                                   \\
RBF       & Radial Basis Function                          \\
ARIMAX    & Average with Exogenous inputs                  \\
DTW       & Dynamic Time Warping                           \\
PLF       & Probabilistic Load Forecasting                 \\
RF        & Random Forest                                  \\
DLF       & Deterministic Load Forecast                    \\
PMP       & Probabilistic Forecasting Model Pool           \\
DMP       & Deterministic Forecasting Model Pool           \\
ANFIS     & Adaptive Neuro-Fuzzy Inference System          \\
DRNN-LSTM & Deep RNN with LSTM                             \\
PSO       & Particle Swarm Optimization                    \\
SVM       & Support Vector Machine                         \\
ESS       & Energy Storage Systems                         \\
BDD       & Bad Data Detection                             \\
AC        & Alternating Current                            \\
SE        & State Estimation                               \\
FDI       & False Data Injection                           \\
DPMU      & Distribution-Level PMU                         \\
PMU       & Phasor Measurement Unit                        \\
FDIA      & FDI Attacks                                    \\
PSSE      & Power System State Estimation                  \\
SCADA     & Supervisory Control and Data Acquisition       \\
MMSE      & Minimum Mean-Squared Error                     \\
AMI       & Advanced Metering Infrastructure               \\
NTL       & Non Technical Loss                             \\
IMM       & Intermediate Monitor Meter                     \\
LSE       & Linear System of Equations                     \\
DBN       & Deep Belief Networks                           \\
HET       & Hidden Electricity Theft                       \\
PPETD     & Privacy-Preserving Electricity Theft Detection \\
FCM       & Fuzzy Cognitive Map                            \\
RL        & Reinforcement Learning                         \\
MG        & Micro Grid                                     \\
HVAC      & Heating Ventilation and Air Conditioning       \\
SLAMP     & Supervised Learning Aided Meta Prediction      \\
V2G       & Vehicle to Grid                                \\
V2H       & Vehicle to Home                                \\
QC        & Quantum Computing                              \\
\end{tabular}
\end{table}

\begin{figure*}[h!]
	\centering
	\includegraphics[width=\linewidth]{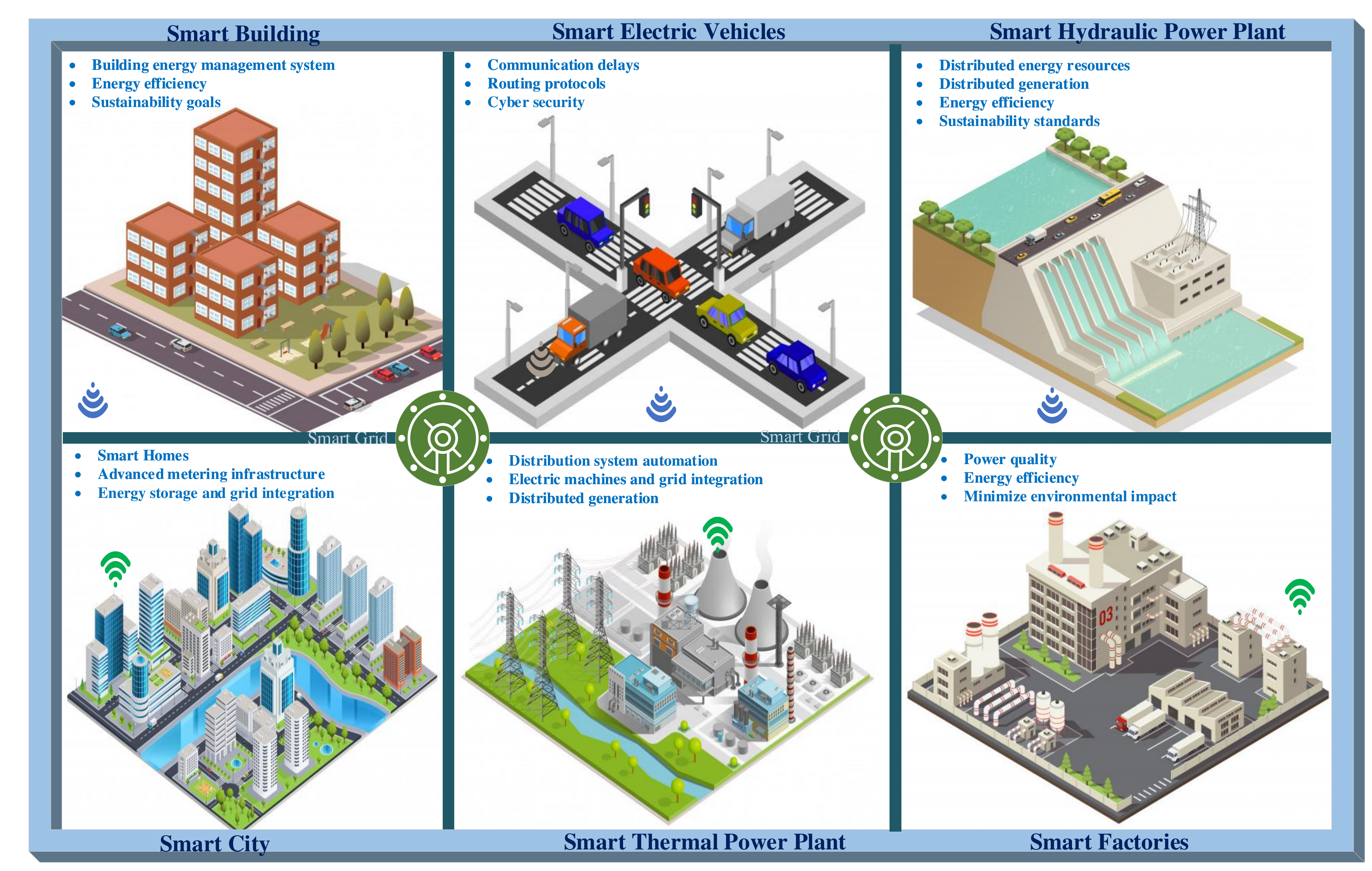}
	\caption{Integration of Smart Grid with other Smart Areas.}
	\label{Fig:Smart Grid}
\end{figure*}

Unlike other commodities like oil, food, etc. electricity cannot be stored for future supply. It has to be dispersed to the consumers immediately as it is produced. The electric grid, in short, grid, is a network of electricity generating stations, transmission lines that carry the electricity to the customers. With the rapid growth of population and increased number of industries, it has become a difficult task for the grids to manage the demand of the electricity for household purposes and industrial purposes. The increased demand of electricity at particular hours of the day lead to several problems like short circuits, failure of transformers. To address these issues of transmission of electricity through traditional grids, there is a necessity to predict the consumption patterns of the customers to effectively deliver the electricity. In this context, the concept of smart grids (SG) has been introduced. A SG can intelligently predict the demands of electricity and hence can transmit the electricity based on the predicted demand. A SG through its intelligent sensing and prediction can address several issues of the traditional grids such as demand forecasting, reduction of power consumption, reducing the risks of short circuits thereby saving the loss of lives and properties \cite{alirezazadeh2020new,kumari2020deal,alazab2020multidirectional}. The advancements in technologies such as the Internet of Things (IoT) \cite{al2015internet}, fifth-generation (5G) wireless networks and beyond \cite{kumari2019fog}, big data analytics \cite{wang2017wireless} and machine learning (ML), have realized the true potential of SGs \cite{deepa2020survey}. SG has several stakeholders and it can be connected with several other smart areas such as smart vehicles, smart buildings, smart power plants, smart city, etc. as depicted in Fig. \ref{Fig:Smart Grid}.

Demand response (DR), one of the most important concepts in SG, encourages the customers to offload non-essential consumption of electricity during the peak hours of the day. DR can effectively manage the supply of electricity in a balanced way during peak hours. By integrating the recent technologies such as ML, IoT and big data analytics with SG, the electricity demand of the customers can be predicted and the DR can be automated \cite{siddiquee2020demand,sarker2020optimal,zhang2020two}. As the amount of data generated from the SG network is vast, deep learning (DL) based models can be used to learn the patterns from the data and predict the demand of electricity and peak hours. 
In the literature, several researchers have investigated the DL concepts for DR in SG. Wen \emph{et al.} \cite{wen2020modified} used a DL model based on recurrent neural networks (RNN) to identify the uncertainties in the SG networks, optimal rates of electricity for each hour, power load. Hong \emph{et al.} \cite{hong2020deep} presented a load forecasting mechanism for residential purposes through DL. Wen \emph{et al.} \cite{wen2020load} proposed a DL based model to predict hourly load demand of electricity for residential buildings in Austin, Texas, USA. Hafeez \emph{et al.} \cite{hafeez2020electric} proposed a factored conditional restricted Boltzmann machine model based on DL to predict hourly electricity load.  Ruan \emph{et al.} \cite{ruan2020neural} proposed a neural-network-based lagrange multiplier selection model to optimize the number of iterations in neural networks for predicting DR in SG. Wang \emph{et al.} \cite{wang2020deep} proposed a DRL method integrated with duelling deep Q network to optimize the DR management in SG. Zhang \emph{et al.} \cite{zhang2020edge} proposed edge-cloud integrated solution for DR in SG using reinforcement learning. 

Many researchers have conducted comprehensive reviews on the applications of DL for addressing security issues, load and price forecasting, energy harvesting, smart metering in SG \cite{cui2020detecting,haque2020machine,sayghe2020survey,huseinovic2020survey,tan2016survey,musleh2019survey,mollah2020blockchain,hu2020modeling,liu2020edge}. 
For example, the reviews on cybersecurity (e.g., false data injection attacks and denial-of-service attacks) in SG systems can be found in \cite{cui2020detecting, haque2020machine,sayghe2020survey,huseinovic2020survey,tan2016survey,musleh2019survey}. 
Mollah \emph{et al.} \cite{mollah2020blockchain} discussed the use of blockchain in addressing various security issues that are caused by the growing number of connections in the centralized SG systems. 
Hu \emph{et al.} \cite{hu2020modeling} presented a contemporary review on various aspects of SG-wireless powered communication (WPC) systems, including utilization, redistribution, trading, and planning of the harvested energy. This work also revealed the application of mathematical tools (e.g., convex and Lyapunov optimization) in optimizing SG-WPC systems along with potential directions in beyond 5G systems. 
We can observe that a comprehensive review on the use of DL for DR in SG is still missing as the existing surveys focused on particular topics such as security and application. 
Moreover, DL has been recently/widely utilized for intelligent solutions in SG and DR systems, but the recent developments may not be included in the existing surveys. Motivated by these observations, in this paper, we provide a comprehensive survey on applications of DL for DR and SG systems. The primary motivation of the current review is to educate the readers on the state-of-the-art on DL, DR, applications of DL in DR and SG. 
    
\subsection{Contributions and Paper Organization}
In a nutshell, the main contributions of this survey are summarized as follows: 
\begin{itemize}
    \item An overview of DL, SG, DR, and the motivation behind the use of DL for SG and DR systems.
    \item A comprehensive review on the applications of DL for SG and DR systems, including electric load forecasting, state estimation, energy theft detection, energy sharing and trading.
    \item Several use cases are presented to illustrate the application of DL in DR and SG systems.
    \item Several challenges, open issues, and future directions on applications of DL for DR an SG are discussed.
    \end{itemize}
     
The rest of the paper is organized as follows. Section~\ref{Sec:Fundamentals} discusses the fundamentals of DL, DR in SG, and motivations behind the use of DL for DR and SG. Section \ref{Sec:Review} presents state-of-the-art on applications of DL in electricity load forecasting, state estimation, energy theft detection, energy sharing and trading in SG. Section \ref{Sec:UseCases} discusses some of the recent use cases for DR and SG. Challenges, open issues, and future directions are presented in section \ref{Sec:Challenges} and the concluding remarks are presented in section \ref{Sec:Conclusion}. For clarity, the list of frequently used acronyms is summarized in Table~\ref{tab:Acronym}.

	
	
	
	
	
	

\section{Background and Motivations}
\label{Sec:Fundamentals}

\subsection{Fundamentals of Deep Learning}
DL has been receiving great attention in ML techniques \cite{bhattacharya2020deep,gadekallu2020deep}. For understanding DL a good knowledge of the ML basics is desirable.
\begin{figure}[t]
	\centering
	\includegraphics[width=\linewidth]{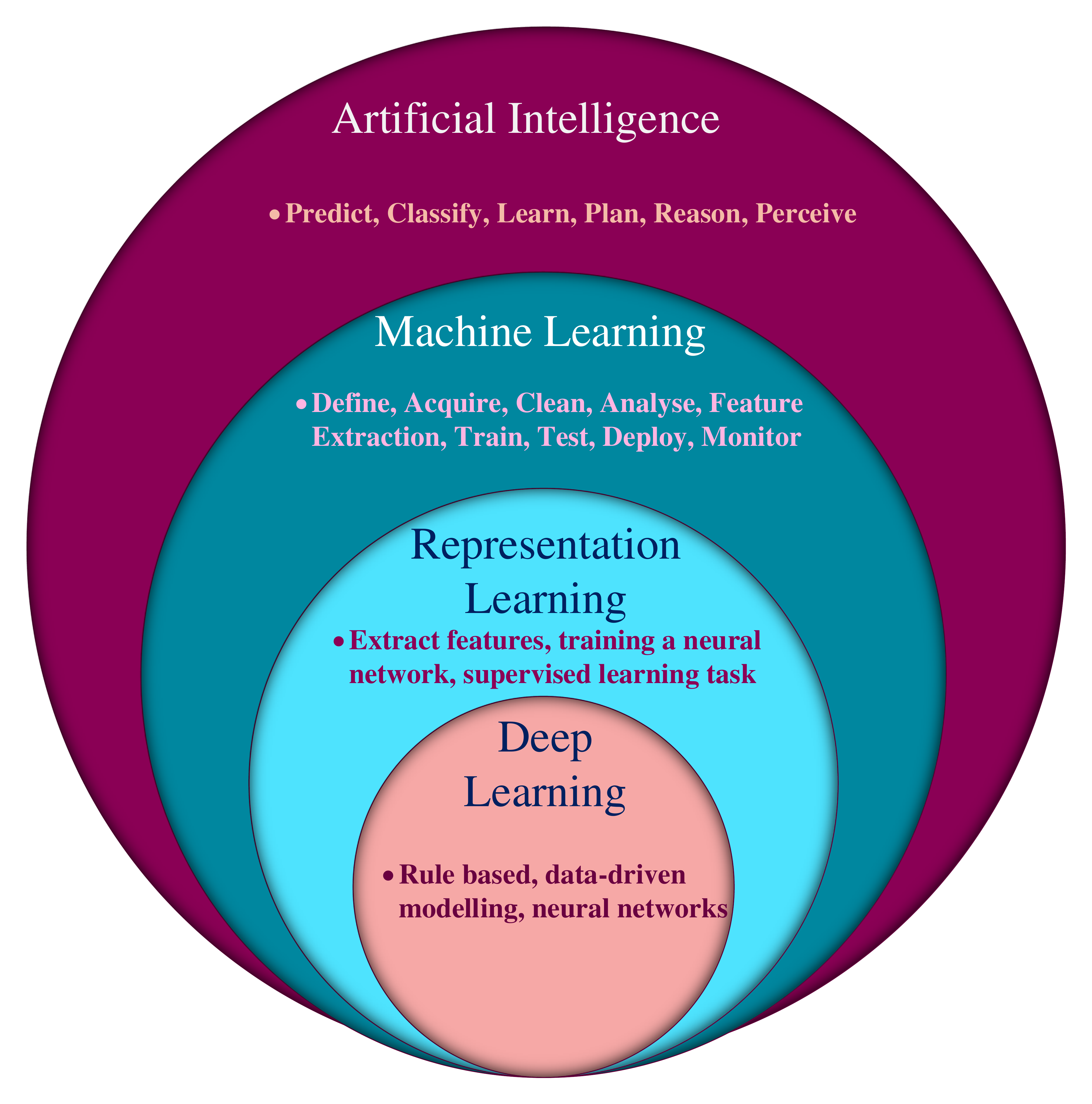}
	\caption{Relation of DL is with RL, ML, and AI.}
	\label{Fig:d1}
\end{figure}
Fig.~\ref{Fig:d1} depicts the relation of DL with representation learning, ML, and artificial intelligence (AI). ML algorithms can learn from data. ML enables us to program the problems that are not solvable with traditional programming techniques. It allows a computer program to learn from experience, concerning a task and evaluates performance measures over the task to improve the learning.
Several tasks can be programmed with ML, such as classification, regression, transcription, sampling, etc. For evaluating the performance of ML algorithms quantitative metrics like accuracy and error rate are used. Based on the experience acquired during the learning process of the algorithm, ML algorithms can be divided into two broad classes on the basis of the learning: supervised and unsupervised. Others are reinforcement learning and recommendation systems.
Supervised learning algorithms learn through a dataset consisting of attributes, whereas each of the evidence is also accompanied by a class label or target class. Linear regression is the fundamental technique of supervised ML.
The main objective of any ML algorithm is to deliver good performance over unknown data not only over training data. The performance of any ML algorithm can be evaluated by its ability to achieve minimum training error  and the difference of training error and generalization error should be small. 
Unsupervised learning algorithms use a dataset consisting of several attributes, then obtain valuable associations from the dataset, for example, clustering algorithms.

Deep neural network (DNN) is the most effectively used technique in DL. DNNs are the extended form of the artificial neural network (ANN) with a significantly higher number of layers to acquire a depth of the network. DL algorithms typically learn the complete probability distribution through the dataset, explicitly as in density approximation, or implicitly as synthesis or denoising. Most DL algorithms are based on stochastic gradient descent optimization algorithms \cite{devan2020efficient}. Usually,  they are a combination of a set of modules, an optimization algorithm, a cost function, a model, and a dataset, to build an ML algorithm.
Present DL facilitates supervised learning by a framework, which includes additional layers and more neurons within them, to design a DNN which can implement today’s increasing complexity requirements. It can map an input vector to an output vector quickly for a given big model and a huge training set. A feed-forward DNN can perform this function effectively, but further regularization, optimization, and scaling the DNN to handle large input vectors, for example, high-resolution image data or heavy length time series data should be specialized. For incorporating these requirements convolutional and RNN were designed.

Convolutional neural network (CNN) is a kind of DNN that employs grid-like topology for processing the data instead of general matrix multiplication in minimum one of the layers. Grid like topology can be adopted, for example, 1-dimensional grid samples at a fixed interval of time in case of time series data, a 2-dimensional grid of pixels for image data. Let $y$ be a measurement at time $t$, we can assume that $y$ and $t$ are defined on integer $t$, the discrete convolution can be represented by the following \cite{N2}:
\begin{equation}
d(t)=(y, w)(t)=\sum_{b=-\infty}^{\infty} y(b) w(t-b).
\end{equation}
In convolution terminology, the function y is an input, b is the age of measurement, $w$ is the kernel and the output $d$ is referred to as a feature map.
Usually, Convolutions are calculated over more than one dimension at a time. Assume, if a two-dimensional data input $R$, a two-dimensional kernel $K$ is used:
\begin{equation}
d(p, q)=(R . K)(p, q)=\sum_{i} \sum_{j} R(i, j) K(p-i, q-j).
\end{equation}

RNNs are a family of DNN, as CNN process grid of values y for an image, an RNN specifically used to process sequence data that contain vector $[\mathrm{y}^{(\mathrm{t})},\dots,\mathrm{y}^{(\mathrm{T})}]$ where $t$ is an index of timestamp ranges from $1$ to $T$. As CNN scales images to variable sizes, similarly RNN scales sequence data to longer sequence variable sizes. A function that includes recurrence can be an RNN. The following equation represents the states of the hidden units in a network.  
\begin{equation}
S^{(t)}=f\left(S^{(t-1)}, y^{(t)} ; \eta\right).
\end{equation}
The network gets trained by employing $S^{(t)}$ as a lossy function of the task-related inputs up to $t$ for mapping an arbitrary length input vector to a fixed length of sequence $S^{(t)}$.
$S$ represents the state at time $t$, function $f(\cdot)$ maps the state at $t$ to the state at $(t+1)$, the same value $\eta$ is used for the next states to parameterized the function $f(\cdot)$ \cite{N2}. 
S. Atef \emph{et al.} \cite{N1} proposed a DL-based prediction with demand and response scheme in the SG to provide an accurate prediction of real-time electricity consumption. The proposed architecture includes to main modules DL module has implemented by a four-layer RNN which predicted hourly electricity consumption patterns by the customers. The second module is the DR decision-making module which uses the predicted consumption patterns from the DL module as input and obtains suitable actions to decrease peak electricity consumption demand for saving cost and energy.

Velasco \emph{et al.} \cite{N3} presented a DNN based power loss prediction model for large scale (LV) supply zones reflecting the inconsistency of the LV supply networks, it constitutes load imbalance, voltage fluctuations from supply generation, and indeterminate smart meter readings. The model employed DNN, the input vector consisting of scaled demand and response data from the entire LV supply region is fed on the input layer I. It was trained using the k-fold-method and obtained the optimal hyper-parameters of the model simultaneously. The output layer O generates the entire technical loss patterns of the LV large scale supply region. The network used multiple hidden layers with n units in each. The model avoided overfitting by dropping out some of the neurons. The output power loss sequence is equated with the target power loss and the mean-squared error is calculated, this error has backpropagated to the previous layer to find the loss function. When the difference between final error and previous error reaches below the threshold error, the network got trained and ready to classify DR patterns.

A generic prediction model of machine tools developed using DL at two stages; Training and utilization \cite{N4}. At the training stage prediction model gets trained using data obtained from the data acquisition system. In the utilization stage, the developed prediction model is exploited for predicting energy usage of machine tools patterns. The training phase is implemented by three steps data acquirement and transformation, unsupervised DL for feature subset selection, and attainment of the energy prediction model. The utilization phase uses the model developed in the training phase to recognize the present energy usage \cite{budhiraja2020energy} of the machine tools by applying the extracted features. 

\begin{figure*}[t]
	\centering
	\includegraphics[width=1.00\linewidth]{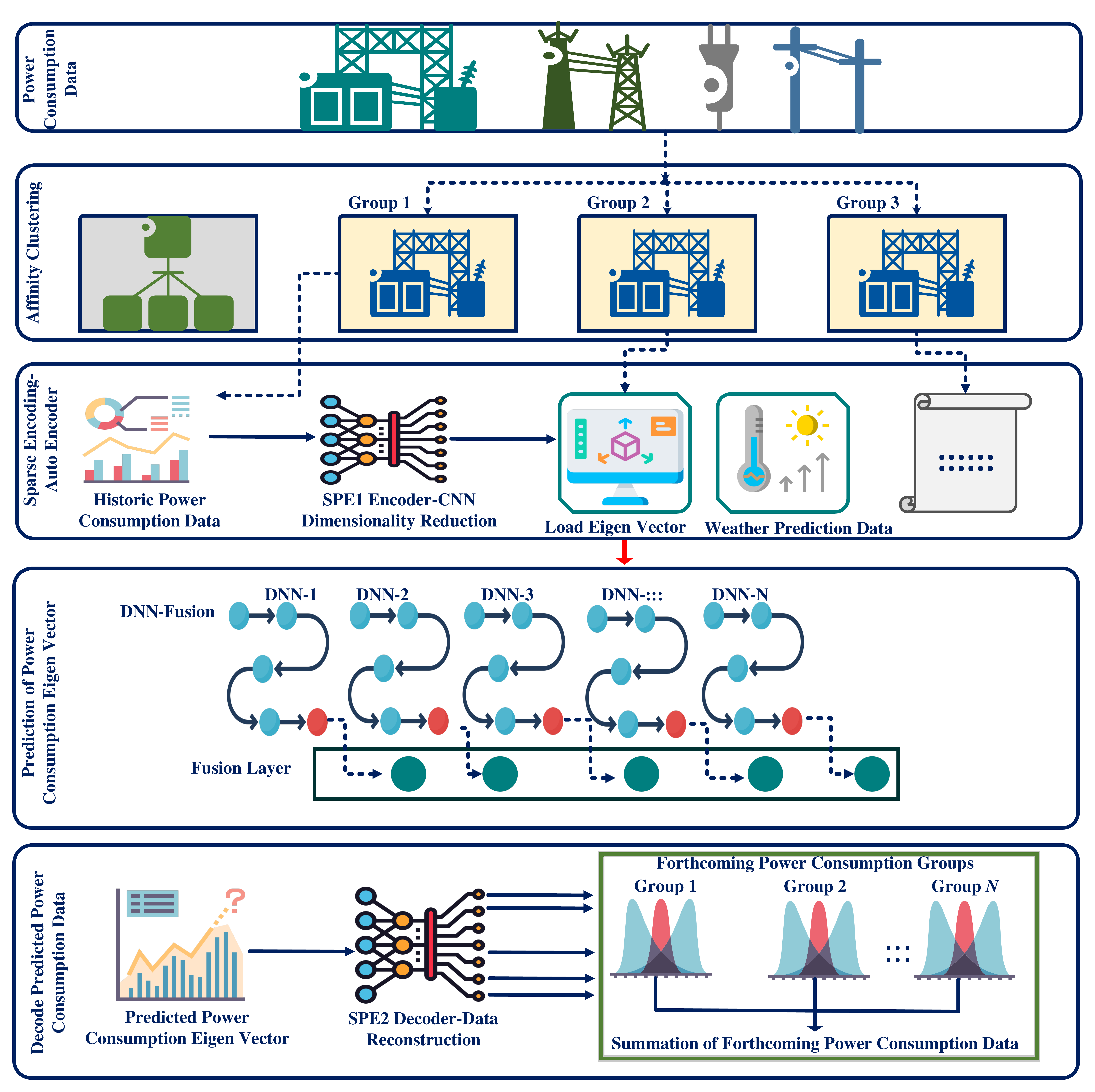}
	\caption{Architecture of the prediction model for aggregated power consumption data.}
	\label{Fig:d2}
\end{figure*}

A day ahead aggregated electric load prediction using two-terminal sparse coding and DNN fusion was developed. The structure of the prediction model shown in Fig.~\ref{Fig:d2} works in 3 steps; first, it splits the aggregated electricity consumption data into clusters by using the Affinity clustering analysis method. Then stored load features are selected by applying sparse auto-encoder. The sparse auto-encoder carry out feature selection using a limited number of units in the hidden layer. After training, the sparse auto-encoder splits into two subunits say SPE1 and SPE2. SPE1 worked as the only encoder to obtain feature selection and SPE2 worked as the only decoder to transform power consumption eigenvector into the load curve. The network design of SPE1 and SPE2 used CNN and long-short term memory (LSTM) respectively. CNN is used to achieve improved accuracy in load prediction and  LSTM is employed to evaluate the power consumption by a single consumer. Further DNNs (fusion of CNN and LSTM) are integrated to achieve the fusion model. Feature-level fusion model is  designed by the following process. 
\begin{itemize}
\item Both the networks are trained independently.
\item The high-level features from the second last layer of both the DNNs are fused to find a new eigenvector.

\item A new completely connected fusion layer is added to obtain a new load eigenvector. Only this layer gets trained in this step. The decoder of auto-encoder generates the forthcoming power graphs of the next day. The final load prediction can be evaluated by adding up the predicted group data \cite{N5}.
\end{itemize}

\begin{figure*}[t]
	\centering
	\includegraphics[width=0.90\linewidth]{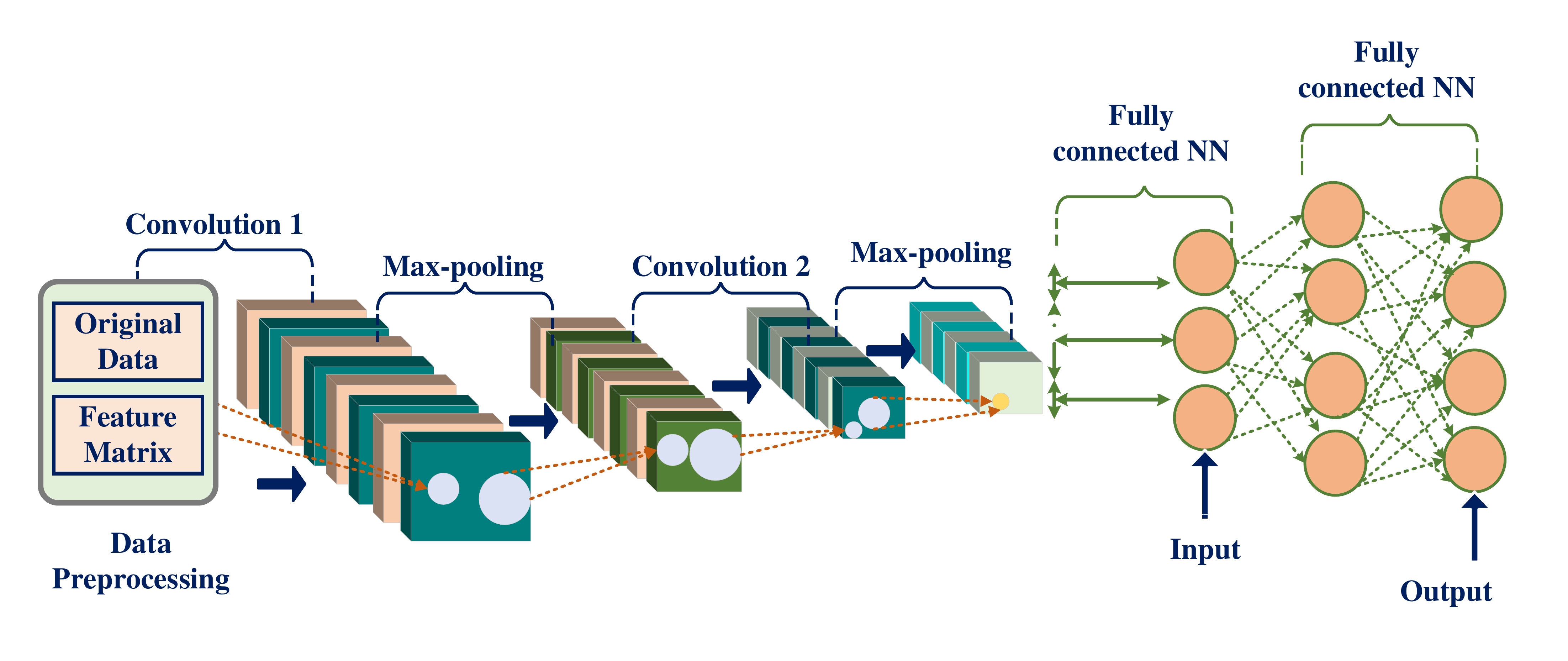}
	\caption{Design structure of the Deep CNN framework.}
	\label{Fig:d3}
\end{figure*}

A huge amount of fine-grained individual consumer load data can be obtained easily through smart meters. \cite{N6} presented a method using DL for the identification of socio-demographic information employing smart meter data. A deep CNN has been used to select features subsets from the electricity consumer-data collected from smart meters. The CNN architecture was constructed using eight layers as shown in Fig.~\ref{Fig:d3}. The first three were convolutional layers, the next three were pooling and one is an entirely connected layer and SVM was the last layer in the architecture. The CNN architecture was determined using two facts consumption patterns of the consumers and the sample in the train set. Reduced features then fed on the pooling layer and lastly a completely connected layer will classify the data. The hyperparameter tuning of CNN was carried out using grid-search with cross-validation techniques. Then SVM employed to extract the behavioral patterns of the consumers automatically. This is used to facilitate the consumers in terms of better service, load efficiency and it also helps in designing SGs.

\subsection{Demand Response in Smart Grids}
To maintain demand and supply trade-off in the load management systems, DR models came into existence. DR is one of the important strategies followed by electricity companies for the reduction of energy consumption by the consumers during peak hours. In this strategy customers can choose to shed non-essential loads by themselves or it can be done automatically by the utility during peak hours of the day \cite{siddiquee2020demand}. DR is an agreement between the customers and the utility based on certain conditions like time intervals, price, and load. Through DR, power consumption can be reduced during the peak hours of the day that results in reduction of operating costs, installation costs, and also the mitigation of potential SG failures. DR leads to the lowering the costs of electricity that can lead to overall reduction of retail prices. Utility companies can encourage DR by offering lower electricity prices during non-peak hours to encourage the customers to shed non-essential loads during peak time.

\subsection{Deep Learning for Demand Response and Smart Grids}
DR models use consumer and their load consumption data for maintaining grid sustainability by load shedding during peak demands and managing the entire consumption data patterns efficiently.
Hence, designing an effective DR system is based on the present time adjustments of load consumption data patterns that need to be extracted to obtain precise load consumption patterns. In order to devise an efficient DR system, an effective prediction technique is to be implemented, so that system can recommend suitable solutions reflecting consumption behaviors by predicting forthcoming consumer usage patterns \cite{N1}.

For SG and smart residential apartment areas, there is a great need for devising a precise energy consumption prediction system. A deep RNN with a gated recurrent unit (GRU) system was developed for predicting energy supply-demand in residential apartments for a small to medium duration of time \cite{N7}. The integrated DRNN-GRU model is a five-layered neural network consisting of optimized hyperparameters with normalized input.  The first layer is an input layer, fed with daily hourly load consumption data samples. The second layer referred to the first GRU layer to produce output for each point time. The third layer referred to the second GRU layer to produce a higher dimension output than the previous layer. This layer has tuned more number of weights and bias. The fourth layer is a simple hidden layer. The fifth layer is an output layer that produces prediction results. The backpropagation is used for parameter optimization, min-max normalization is used to scale the train data. Overfitting issues are addressed by adding regularization to the cost function and dropout was applied to the hidden layer. The model can cooperate with time dependencies of data, it is also effective in solving missing information problems through historical data.

An optimized DR system is presented \cite{N8} for managing interruptible load (IL) under the time-of-use pricelist for flexible electricity load patterns. It employed a DRL technique with dueling deep Q network (DDQN) patterns, which made the realistic usage of DR possible. The IL problem has been implemented employing a Markov decision process to gain the highest profit for a longer duration, which defines the state, operation, and reward function of the DR system. The DDQN  organization achieved reduced noise and enhanced the stability of the model, it shows that the instability in the convergence of DDQN has been reduced to obtain the appropriate value of the loss function. Finally, the model realized the reduction in the peak electricity consumption demand and the cost of operation of the regulation of voltage within the limit without compromising safety. 

The problem of heavy electricity consumption in industry sectors is addressed by Renhzi Lu \emph{et al.} \cite{N9}. A model is developed using multiagent DRL for implementing DR strategy for managing a distinct manufacturing system. The industrial manufacturing system has been formulated as POMG initially, later, the MADDPG algorithm was applied and an optimal schedule of load consumption for every machine has been designed. Then the model was deployed on a battery manufacturing system for evaluating its effectiveness. The results of the simulation proved that the DR system could provide the least cost of power consumption and maintained fewer production expenditures in comparison to the non-DR benchmark system. The model enables the system to get reduced electricity consumption and instability of the grid. 

An efficient STLF framework is developed to address the challenges presented by vigorous and stochastic consumer behavioral patterns in residential campuses. The framework predicts individuals’ electricity consumption patterns using aggregated load data of appliances and correlation amongst them by employing DL techniques. The framework comprises data acquisition, data preprocessing, learning, and load prediction modules. The correlation among various load data in appliances is modeled by the multiple time series method and included in the framework. The learning is applied through the DNN and ResBlock techniques. A grid search algorithm is utilized for hyperparameter tuning in DNN \cite{N10}.

Yandong Yand \emph{et al.} \cite{N11} presented a deep ensemble-learning with a probabilistic model for energy prediction in an SG for reliable energy consumption management. The model can address the problem of determining the accurate power requirement of each consumer with an efficient SG. Existing methods are not able to cope up with the aggregated load profiles of consumers, irregular load patterns, and instability. The ensemble load prediction model is implemented on different consumer group profiles. The model is designed to work in four phases; representation learning, clustering of consumer profiles, multiple task feature learning on the clusters; where the number of tasks is equal to the number of clusters formed, and deep ensemble learning; DNNs are ensembled for probabilistic prediction. The model performed initialization and batch training in a randomized manner and made the scalable and variable deep ensembling framework by applying different kinds of DNNs, which is preferable for distributed and parallel computing environments. Lastly, the generated prediction results are passed through the LASSO-based quantile technique to get polished ensemble predicted patterns. The framework is deployed on SME consumers to demonstrate the effectiveness and improvement of the existing systems. It can be employed on the DR and electricity requirement management system for homes in a SG.

From the above discussion it is evident that due to enormous quantity of the data generated through sensors and other IoT devices in SG, traditional ML algorithms are incapable of extracting trends, patterns and predictions. DR plays a vital role in reducing the load of the SGs during peak hours. To effectively extract patterns from the SG and the customers' electricity usage patterns to automate shedding of unnecessary load during peak hours, DL based models are ideal solutions due to their scalability. Apart from electricity usage patterns of customers DL can play a very important role in uncovering electricity thefts, tampering of the meters, predict fire accidents, etc. in a SG. These benefits of DL in SG and DR is the main motivation behind the current review paper.

\section{State-of-the-Art DL Approaches for Demand Response and Smart Grids}
\label{Sec:Review}

In this section several state-of-the-art works on applications of DL for DR and SG like electric load forecasting, state estimation in SG, energy theft detection in SG, energy sharing and trading in SG are discussed.

\subsection{Electric Load Forecasting}


Load forecasting is the technique of predicting the power or energy required to meet the future demand. It is of high significance in the energy and power sectors.  Accurate load forecasting is necessary for the effective planning and operation of power systems. It can be divided into two categories namely short-term load forecasting and long-term load forecasting.

Typically, in short-term load forecasting (STLF), load  for the next-hour up to the next two weeks is estimated. \cite{zhu2019novel} considered forecasting the Electric Vehicles (EV) charging load as their large penetration lead to high uncertainty in power demand of the power system. The paper uses ANN and LSTM based DL approaches for plug-in EVs load
forecasting. A charging station company of PEVs is used to compare both traditional ANN approach and LSTM approach. It was observed that the LSTM model had lower errors and higher accuracy in short-term EVs load forecasting.

The authors in \cite{aly2020proposed} proposed six different models which involve combinations of ANN, wavelet neural network, and kalman filtering  schemes for STLF. These models were validated on different data sets and it was seen that compared to the conventional models in the literature, all six models had shown higher accuracy. \cite{tang2019ensemble} studied the large timespan
quasi-periodicity of load sequences and came up with an ensemble method combining Auto-regressive Integrated Moving Average (ARIMA) and LSTM. It was evaluated on a dataset of 737 weeks of load consumption and the superior performance of this model was observed when compared with the popular STLF models.

The extensive use of distributed generations in smart-grids brought additional need for the accuracy of STLF. To address this, a method based on rolling update (RU), attention mechanism (AM), and bi-directional
long short-term memory (Bi-LSTM) was proposed by \cite{wang2019bi}. Its validity is measured on the actual data sets from different countries and it was not only proved that this method had higher accuracy but also it required less computation time compared to other models. To further improve the STLF, the authors in \cite{khwaja2020joint} has given a joint bagged-boosted ANN model which combines both bagging and boosting while training. It consists of training ensembles of ANNs in parallel where each ensemble consists of training ANNs in sequence. Statistical analysis on real data showed that this model had lower bias, variance and forecasting error than single ANN, bagged ANN and boosted ANN.

There was an idea to use multiple kernel learning (MKL) for residential short-term electric load forecasting than the traditional kernels as it might invoke more flexibility. But using conventional methods create complex optimization problems. To solve this issue, Wu \emph{et al.} \cite{wu2019multiple} proposed a gradient-boosting based MKL which used a boosting-style approach. It also had less computation time. In addition to this method, the paper also proposed transfer based learning algorithms so that the learned information from source houses can be transferred to target houses. The results showed that these transfer algorithms reduced the forecasting error when only limited data was available.

The authors in \cite{dagdougui2019neural} presented a study for hour-ahead and day-ahead load forecasting in a district building using ANN. It evaluated these ANNs using Levenberg-Marquardt (LM) and bayesian regularization (BR) as the learning algorithms. After analysing the model on different buildings with respect to both day-ahead and hour-ahead forecasting, it was seen that hour-ahead predictions yield comparatively better performance. This paper also gave the best strategies for better performance of day-ahead predictions so that electricity bills can be reduced.

Even though the recent RNNs have a good performance in load forecasting, they do not use a predicted future hidden state vector or fully use the past information. If any errors are present in the hidden state vector, it cannot be corrected by existing RNN models so that better forecasting in the future is hindered. To address this issue, the authors in \cite{kim2019recurrent} proposed a recurrent inception CNN which was a combination of RNN and 1D-CNN. Experiments showed a better performance of this model than multi-layer perceptron, RNN, and 1D-CNN in daily electric load forecasting.

Due to the high computation complexity of support vector regression (SVR) approach in short-term
electric load forecasting, a sequential grid approach based SVR is proposed by  \cite{yang2019sequential} so that the estimation becomes more efficient and accurate. The experiments for short-term forecasts demonstrate the better performance and accuracy of this model compared to regular SVR model. There is a limitation of the paper that it does not consider the effect of interaction between forecasting and subsampling. For the same short-term electric load forecasting, \cite{hu2019short} presented a model based on the hybrid particle swarm optimization-genetic algorithm-back propagation neural networks (PSO-GA-BPNN) algorithm. Here the PSO-GA algorithm is used in optimizing BPNN parameters. To analyze this model, the data is taken from different papermaking enterprises and it is seen that this model is superior to the PSO-BPNN and GA-BPNN based models.

To improve the accuracy of STLF, a proper load data analysis and more effective feature selection are considered by \cite{tang2019application}. The load data is first made into clusters by  Bisecting K-Means Algorithm and then it is decomposed into intrinsic mode functions by ensemble empirical mode decomposition. Then the paper proposed a forecasting model based on bidirectional RNNs (Bi-RNN) and deep belief network (DBN). It was verified on a power grid load data and it was proved to be better than other methods. DBN is also used in \cite{ouyang2019modeling} to predict the hourly load of the power system. Before that, this paper uses Copula models to compute the peak load indicative variables. Experiments prove that this framework is superior to the classical neural networks, SVR machine, extreme learning machine, and classical DBNs in short term power load forecasting.

A conditional hidden semi-Markov model is used by \cite{ji2019data} for load modeling of residential appliances. And then a load-prediction algorithm was given using the learned model. This model is evaluated using 1-minute resolution data from different appliances and it is found that this model is effective and applicable. Also, in the residential appliances energy disaggregation is one of the main problems, especially in the type II appliances which serve multiple purposes. \cite{kong2019practical} focused on this issue and proposed a deep convolutional neural network (DCNN) based framework and a post-processing method to solve this non-intrusive load monitoring issue. Experiments are conducted on a public dataset and it is observed that this model has a better performance over other works.

The authors in \cite{sun2019using} proposed a novel probabilistic residential net load forecasting model to capture the uncertainties caused by distributed photovoltaic (PV) generation. It uses a Bayesian deep LSTM neural network. Experiments conducted on the data from the Australian grid prove that the proposed method is superior to some state-of-the-art methods. The paper also considered  clustering in subprofiles and PV visibility which are major contributors to the performance of the model.

A fuzzy-based ensemble model is presented in \cite{sideratos2020novel} for week-ahead load forecasting. This model uses hybrid DL neural networks combining ANNs, ensemble forecasting. A fuzzy clustering idea is used to divide the input into clusters and then these are trained to a neural network consisting of one pooling, one convolutional, one radial basis function (RBF), and two fully-connected layers. It is tested on two case-studies and proved to be more effective than traditional models. The authors in \cite{chen2019day} considered a model based on fusion of DNN and two-terminal sparse coding to forecast day-ahead aggregated load. The challenges like redundant features caused by high-dimensionality are overcome by this two-terminal sparse coding. Clustering is used to to group the load into different groups and different structured DNNs are used in predicting load of a single group. Then these are fused using a fusion layer. The effectiveness of this model is demonstrated by case-studies.

As DL models require high training costs like time and energy, it becomes difficult to apply these models in the real world. Hence, to minimise the training cost without compromising on accuracy,  the authors in \cite{huang2019loadcnn} proposed a model based on CNN, called LoadCNN. It is used in forecasting day-ahead residential load with minimizing training costs. Experiments show that the training costs (time, carbon-dioxide emissions) are around 2\% of the other state-of-the-art models and also the prediction accuracy is equal to the current models.

One of the traditional load forecasting methods is autoregressive integrated moving average with exogenous inputs (ARIMAX) which is a time-series forecasting method. Cai \emph{et al.} \cite{cai2019day} proposed RNN and CNN formulated under direct multi-step and recursive manners. Their accuracy, efficiency are compared with the traditional ARIMAX. The proposed model proved to have better performance with accuracy improved by 22.6\% compared to ARIMAX.  The authors in \cite{gasparin2019deep} focus on feed forward neural networks, sequence to sequence (seq2seq) models, temporal CNNs and RNNs in electric load forecasting. Based on two datasets, the above models are analysed. It is observed that Elmann RNNs perform better compared to GRU and LSTM with less cost in aggregated load forecasting. Also temporal CNNs have shown a good performance and are also observed to have the potential to introduce an advance in the future developments of power systems.

A model based on dynamic time warping (DTW) distance and gated RNNs is proposed by Yu \emph{et al.} \cite{yu2019deep} for daily peak load forecasting.The shape-based DTW distance captures the changes in load, and a three-layered gated RNN is used in forecasting the load. Simulations are performed on the EUNITE dataset. The proposed model is proved to outperform the state-of-the-art forecasting methods.

Probabilistic load forecasting (PLF) which provides forecasts as a probability density function or prediction intervals becomes critical when it comes to forecasting unpredictable and volatile load.  Yang \emph{et al.} \cite{yang2019bayesian} proposed a Bayesian DL-based multitask PLF framework for forecasting residential load. The model follows a three stage pipeline where clustering, pooling and multitask learning are employed at each stage respectively. Results show that overfitting issue is addressed by this
strategy. Also better performance of this model was observed compared to conventional methods like SVR, pooling-based LSTM and random forests (RF). Wang \emph{et al.} \cite{wang2019probabilistic} proposed a pinball loss guided LSTM model for residential PLF. Here, pinball loss was used in training the parameters, instead of mean square error. This enabled the traditional LSTM point-based model to extend to probabilistic forecasting. The results obtained demonstrated the effectiveness of this model.

Fand \emph{et al.} \cite{feng2019reinforced} developed dynamic model selection based on a STLF model using Q-learning. It gives both deterministic load forecasts (DLF) and PLF. Firstly, 4 predictive distribution models and 10 state-of-the-art ML DLF models are used to form a probabilistic forecasting model pool (PMP) and a deterministic forecasting model pool (DMP). Then a 2-step process happens in which a Q-learning agent chooses the optimal DLF and PLF models from PMP and DMP respectively. Experiments on smart meter and two-year weather data showed that this model reduces the PLF and DLF errors by 60\% and 50\% respectively compared to other traditional models.

Long-term load forecasting refers to estimating or predicting the load for the next few months or years. It is an important factor for the future planning, operation, and expansion of a power system. Dong \emph{et al.} \cite{dong2019hybrid} presented a model to forecast the annual load of distribution feeders. It uses LSTM and GRU models to exploit patterns in multi-year data. Using the data of an urban grid, LSTM and GRU networks are compared with ARIMA, bottom-up, and feed-forward neural networks. Results demonstrate the superior performance of this model and also GRU model is observed to be faster than LSTM model.

For accurate long-term forecasts based on land use plans, temporal LF and a data-driven bottom-up spatial approach is presented in \cite{ye2019data}. It uses land plots as the basic LF resolution. Also, to aggregate load densities, kernel density estimation and adaptive
k-means are used. Case studies demonstrate that this method is feasible to handle big data and the presented SLF model is more applicable than the benchmark methods.

Mohammad \emph{et al.} \cite{mohammad2019energy} presented a DNN based model to forecast the energy consumption in SG. Deep feed-forward neural network and deep RNN are investigated in this paper. Simulations on the NYISO dataset proved the superiority of the proposed model. The authors in \cite{chapaloglou2019smart} provided an algorithm based on feed-forward ANNs for power flow management of an island's power system to forecast short term day ahead load. It is then passed to a pattern recognition algorithm and based on the curve shape of the load, it is classified. Results of dynamic simulations demonstrated that a combined effect of peak shaving with renewable energy and smoother diesel generator operation is achievable by the proposed algorithm.

Motepe \emph{et al.} \cite{motepe2019improving} has proposed a hybrid DL and AI model for distribution network load forecasting in South Africa. The state-of-the-art hybrid DL technique and a AI technique investigated are LSTM and optimally pruned extreme learning machines, respectively. These are compared with adaptive neuro-fuzzy inference system (ANFIS) and LSTM was seen to achieve higher accuracy. To forecast the PV power output and residential power load, a deep RNN with LSTM (DRNN-LSTM) model is proposed by \cite{wen2019optimal}. To optimize the load dispatch of grid-connected community microgrid, particle swarm optimization (PSO) algorithm is used. 
Results on two real-world data sets prove that the DRNN-LSTM model outperforms the support vector machine (SVM) and multilayer perceptron network. It is also observed that load dispatch optimization enables EVs, and energy storage systems (ESS) to shift the peak load and reduce 8.97\% of the daily costs.
We summarize the existing works on electric load forecasting in Table \ref{tab3.1}. 

\begin{table*}[ht!]
\centering
\caption{Summary on Electric Load Forecasting. }
\label{tab3.1}
\resizebox{\textwidth}{!}{%
\begin{tabular}{|l|p{4.7 cm}|p{2.8 cm}|p{7.35cm}|}
\hline
Ref. &
  Contribution &
  DL Approaches &
  Key features \\ \hline
\cite{zhu2019novel} &
  Short-term EVs load forecasting &
  ANN and LSTM &
  Forecasting the EVs charging load as their large penetration lead to high uncertainty in power demand of the power system \\ \hline
\cite{tang2019ensemble} &
  Large timespan quasi-periodicity of load sequences &
  ARIMA and LSTM &
  Evaluated on a dataset of 737 weeks of load consumption and the superior performance of this model was observed when compared with the popular STLF models \\ \hline
\cite{wang2019bi} &
  Validity is measured on the actual data sets from different countries &
  AM, RU and Bi-LSTM &
  The extensive use of distributed generations in smart-grids brought additional need for the accuracy of STLF \\ \hline
\cite{khwaja2020joint} &
  Joint bagged-boosted ANN model &
  Training ensembles of ANNs &
  Statistical analysis on real data showed that this model had lower bias, variance and forecasting error than single ANN, bagged ANN and boosted ANN \\ \hline
\cite{dagdougui2019neural} &
  Day-ahead and hour-ahead load forecasting district building &
  ANN, BR and LM &
  Best strategies for better performance of day-ahead predictions so that electricity bills can be reduced \\ \hline
\cite{kim2019recurrent} &
  Predicted future hidden state vector &
  RNN and 1D-CNN &
  If any errors are present in the hidden state vector, it cannot be corrected by existing RNN models so that better forecasting in the future is hindered \\ \hline
\cite{yang2019sequential} &
  Sequential grid approach based SVR &
  Support vector regression &
  Due to the high computation complexity of SVR approach in short-term electric load forecasting, a sequential grid approach based SVR  becomes more efficient and accurate \\ \hline
\cite{hu2019short} &
  Consider the effect of interaction between forecasting and subsampling &
  GA-PSO-BPNN &
  GA-PSO algorithm is used in optimizing BPNN parameters, the data is taken from different papermaking enterprises \\ \hline
\cite{tang2019application} &
  Load data analysis and more effective feature selection &
  Bi-RNN, DBN &
  To improve the accuracy of STLF, a proper load data analysis and more effective feature selection \\ \hline
\cite{ouyang2019modeling} &
  Predict the hourly load of the power system &
  DBN &
  Uses Copula models to compute the peak load indicative variables \\ \hline
\cite{kong2019practical} &
  Solve non-intrusive load monitoring issue &
  DCNN &
  In the residential appliances energy disaggregation is one of the main problems, especially in the type II appliances which serve multiple purposes \\ \hline
\cite{sun2019using} &
  Clustering in subprofiles and PV visibility are major contributors to the performance of the model &
  Bayesian deep LSTM neural network &
  A novel probabilistic residential net load forecasting model to capture the uncertainties caused by distributed PV generation \\ \hline
\cite{sideratos2020novel} &
  A fuzzy-based ensemble model is presented for week-ahead load forecasting &
  Hybrid DL neural networks &
  A fuzzy clustering idea is used to divide the input into clusters and then these are trained to a neural network consisting of one RBF, one convolutional, one pooling and two fully-connected layers \\ \hline
\cite{chen2019day} &
  Two-terminal sparse coding and DNN fusion for day-ahead aggregated load forecasting &
  DNN &
  The challenges caused by high-dimensional data such as redundant features are overcome by this two-terminal sparse coding \\ \hline
\cite{huang2019loadcnn} &
  Forecasting day-ahead residential load with minimizing training costs &
  LoadCNN &
  To minimise the training cost without compromising on accuracy \\ \hline
\cite{cai2019day} &
  RNN and CNN are formulated under recursive and direct multi-step manners &
  RNN and CNN &
  Accuracy, efficiency are compared with the seasonal ARIMAX and gated 24-h CNN performed in direct multi-step manner is proved to have best performance with accuracy \\ \hline
\cite{gasparin2019deep} &
  Focused on feed forward neural networks,seq2seq models, temporal CNNs and RNNs in electric load forecasting &
  Temporal CNNs and RNNs &
  Elmann RNNs perform comparably to GRU and LSTM with less cost in aggregated load forecasting \\ \hline
\cite{yang2019bayesian} &
  Proposed a Bayesian DL-based multitask PLF framework for forecasting residential load &
  Bayesian DL &
  The model follows a three stage pipeline where clustering, pooling and multitask learning are employed at each stage respectively \\ \hline
\cite{dong2019hybrid} &
  To forecast the annual load of distribution feeders &
  LSTM and GRU &
  Uses advanced sequence prediction models LSTM and GRU to exploit the sequential information hidden in multi-year data \\ \hline
\cite{chapaloglou2019smart} &
  Power flow management of an island's power system &
  Feed-Forward ANNs &
  Uses a model based on feed-forward ANNs for short term day ahead load forecasting \\ \hline
\cite{wen2019optimal} &
  Uses PSO algorithm to optimize the load dispatch of grid-connected community microgrid &
  DRNN-LSTM &
  Load dispatch optimization enables EVs, and ESS to shift the peak load and reduce 8.97\% of the daily costs \\ \hline
\end{tabular}%
}
\end{table*}


\subsection{State Estimation in Smart Grids}
Estimation of state variables is necessary in order to have a complete, real time and accurate solution for any power system. State variables include voltages, angles, etc. Inputs to the state estimators are just these imperfectly defined state variables. State estimation suppresses faulty measurements and produces the best estimate for any particular power system. The online state-of-health estimation shows low accuracy and also affects the accuracy level of state-of-charge estimation. In order to overcome one of such low accuracy issues of state variables, Song \emph{et al.} \cite{song2020hybrid} demonstrated a joint lithium-ion battery state estimation approach which makes use of data-driven least-square SVM and model-based unscented-particle-filter. The direct mapping models involved omit identification of parameters and updation in complex operating conditions. This correction of state-of-health in state-of-charge estimation gave rise to the joint estimation with different time scales. Experiments revealed that the estimated error of maximum state-of-charge life cycle is not more than 2\% and the state-of-health's root-mean-square-error estimation is not more than 4\%. Results of the implemented model by the authors showed that both state-of-health and state-of-charge can be evaluated with notable accuracy using this demonstrated hybrid estimation procedure.
Dynamic state estimation is generally used to give real-time and proper management for the operation of SG. 

Chen \emph{et al.} \cite{chen2019novel} proposed a new online based method to detect data injection cyber attacks on dynamic state estimation in SG. The conventional data injection attack properly escaped the bad data detection (BDD). In this work, an imperfect data injection attack directed at kalman filter's state variables is considered. They designed a new methodology to select these targeted state variables and then by solving an ideal model related to the PSO algorithm, their values determined. Numerical experiments by the authors show the robustness of the proposed detection method and it is also possible to get the complex attack mechanisms of various attacks. This work can be extended by collaborating with various cyber-attacks and perform feature modeling. 

ANN gives notable accuracy for nonlinear alternating current (AC) state estimation (SE) in SG over conventional methods. But, research claimed that some adversarial examples can trick ANN easily. Liu \emph{et al.} \cite{liu2019adversial} presented a study on  adversarial false data injection (FDI) attack against AC SE with ANN: by injecting a conscious attack vector into measurements, while remaining undetected the accuracy of ANN SE can be declined by the attacker. To create attack vectors, the authors proposed a gradient-based and population-based algorithm. The productiveness of algorithms implemented are checked using simulations on IEEE 9-bus, 30-bus and 14-bus systems under multiple attack layouts. The results from simulations confirm that DE is more effective than SLSQP on all simulation scenarios. The examples of attacks that use the DE algorithm properly declined the accuracy of ANN SE at a very notable probability.

Zhou \emph{et al.} \cite{zhou2020bayesian} addressed and studied the challenges of estimating the distribution of harmonic voltages and locating harmonic sources using unbalanced 3-phase power distribution systems. They developed a model for harmonic state estimation that uses two different types of measurements from distribution-level PMUs (DPMUs) and smart meters. This model needs fewer DPMUs than nodes which makes it a notable application to distribution grids. The authors showed the robustness of the implemented model using immense numerical simulations on an IEEE test feeder. They also probed how the performance of their state estimator could get affected by distributed energy resources' increased penetration.

Mestav \emph{et al.} \cite{mestav2019bayesian} addressed the challenges of state estimation for unobservable distribution systems. For real-time applications, they implemented a model to Bayesian state estimation using a DL approach. This procedure is made up of a Monte Carlo technique in order to train a DNN for state estimation, distribution learning of stochastic power injection, and a Bayesian bad-data detection and filtering algorithm. Experimental simulations showed the effectiveness of Bayesian state estimation for unobservable systems. Direct Bayesian state estimation using DL neural networks surpasses the conventional benchmarks when compared to pseudo-measurement techniques.

Along with an increase in phasor measurement units (PMU) that is being used by utilities for proper and safe tracking of power systems, risk of various cyber-attacks is also increasing. The authors in \cite{basumallik2019packet} studied FDI attacks (FDIA) that aims to change PMU measurements which results in false state estimation. They extracted multi-variate time-series signals from PMU data packets aggregated in phasor data concentrators related to various events such as generation and load fluctuations, line faults and trips, shunt disconnections and FDIA prior to every cycle of SE. To check and prove the PMU data, they proposed a CNN data filter with categorical cross entropy loss and Nesterov Adam gradient descent. This CNN-based filter showed notable accuracy compared to all other classifiers. All experimental simulations were done on IEEE-118 bus and IEEE-30 bus systems.

The schemes such as power system state estimation (PSSE) have become more expensive with emerging applications. To overcome such challenges, Zhang \emph{et al.} \cite{zhang2019realtime} proposed a DNN based model for tracking power systems. On making an iterative solver that was actually made use of an exact ac model, a new model-specific DNN was implemented for real-time PSSE that needs only reduced tuning effort and  offline training. Numerical tests revealed the superiority of the proposed DNN-based approaches when compared to conventional methods. Experimental results on the IEEE 118-bus prove that the DNN-based PSSE scheme performs better than the existing methods like widely used Gauss–Newton PSSE solver.
When there is no noise, PSSE is just equal to solving a system of quadratic equations which also corresponds to analysis of power flow is NP-hard in general. 

Following assumptions like the availability of all power flow and voltage magnitude measurements, Wang \emph{et al.} \cite{wang2019robust} suggest a general algebraic method to convert the power flows into rank-one measurements. They developed a perfect proximal-linear algorithm based on composite optimization. They also implemented conditions on l1 based loss function such that perfect recovery and quadratic convergence of the proposed method are promised. Experimentally simulated tests using various IEEE benchmark systems under multiple settings supported their theoretical developments, also the effectiveness of their methods.

The authors in \cite{mestav2019state} addressed the challenges of distribution system state estimation using smart meters and reduced supervisory control and data acquisition (SCADA) measurement units. To overcome the hurdle of limited measurements, they  determined a Bayesian state estimator that uses DL. This method consists of two steps. Firstly, a deep generative adversarial network is trained to learn the distribution of net power injections at the loads. Then, to obtain minimum mean-squared error (MMSE) estimate of the system state, a deep regression network is trained on samples obtained from the generative network. The simulation results showed that the accuracy as well as the online computation cost of the method are notable when compared to already existing methodologies. 

\begin{table*}[ht!]
\centering
\caption{Summary on State Estimation in Smart Grids.}
\label{tab3.2}
\resizebox{\textwidth}{!}{%
\begin{tabular}{|l|p{5.2 cm}|p{2.3 cm}|p{8.0 cm}|}
\hline
Ref. &
  Contribution &
  Approaches &
  Key features \\ \hline
\cite{song2020hybrid} &
  Data driven least square SVM and model based unscented particle filter &
  SVM, Unscented particle filter &
  The direct mapping models involved omit identification of parameters and updation in complex operating conditions \\ \hline
\cite{chen2019novel} &
  Data injection cyber attacks against dynamic state estimation in SG &
  Kalman filter &
  Conventional data injection attack properly escaped BDD whereas an imperfect data injection attack which is directed at state variables against Kalman filter estimation is considered \\ \hline
\cite{liu2019adversial} &
  Adversarial FDI attack against AC, SE with ANN &
  ANN &
  Proposed a population-based algorithm and a gradient-based algorithm \\ \hline
\cite{zhou2020bayesian} &
  Developed a model for harmonic state estimation &
  DPMUs &
  Addressed the challenges of locating harmonic sources and estimating the distribution of harmonic voltages using unbalanced three-phase power distribution systems \\ \hline
\cite{mestav2019bayesian} &
  Bayesian state estimation using a DL approach &
  DNN &
  Direct Bayesian state estimation using DL neural networks surpasses the conventional benchmarks when compared to pseudo-measurement techniques \\ \hline
\cite{basumallik2019packet} &
  FDIA aims to change PMU measurements resulting in false state estimation solutions &
  CNN &
  Extracted multi-variate time-series signals from PMU data packets aggregated in phasor data concentrators related to various events such as line faults and trips, generation and load fluctuations, shunt disconnections and FDIA prior to every cycle of state estimation. \\ \hline
\cite{zhang2019realtime} &
  A new model-specific DNN was implemented for real-time PSSE that needs only offline training and reduced tuning effort &
  DNN &
  When there is no noise, PSSE is just equal to solving a system of quadratic equations which also corresponds to analysis of power flow is NP-hard in general \\ \hline
\cite{wang2019robust} &
  A general algebraic method to convert the power flows into rank-one measurements &
  Proximal-linear algorithm &
  Developed a perfect proximal-linear algorithm based on composite optimization, implemented conditions on l1 based loss function such that perfect recovery and quadratic convergence of the proposed method are promised \\ \hline
\cite{mestav2019state} &
  To overcome the hurdle of limited measurements they determined a Bayesian state estimator that uses DL &
  SCADA &
  To learn the distribution of net power injections at the loads, a deep generative adversarial network is trained, this is the first step. Then, using the samples obtained from the generative network to obtain MMSE estimate of the system state, a deep regression network is trained \\ \hline
\cite{matthiss2019using} &
  To estimate the grid state, pseudo-measurements based on standard load profiles along with PV generation estimates are used &
  Near-optimal placement &
  Used various meter placement schemes with the aim to improve the estimation accuracy with less investment to a German distribution grid \\ \hline
\end{tabular}%
}
\end{table*}

To estimate the grid state, most of the time pseudo-measurements from PV-generation estimates and standard-load-profiles are used. If one wants to increase the accuracy of the grid state estimation, more measurement units can be introduced into the grid. A near-optimal placement of the measurements is vital for cost-effective operation of the grid. The authors in \cite{matthiss2019using} used various meter placement schemes to reduce investments and improve the estimation accuracy on a German distribution grid. The effectiveness of the placement schemes on the estimation accuracy are compared and benchmarked with respect to the computational efficiency in this study. We summarize the existing works on state estimation in SG in Table~\ref{tab3.2}. 

\begin{figure*}[t]
	\centering
	\includegraphics[width=1.0\linewidth]{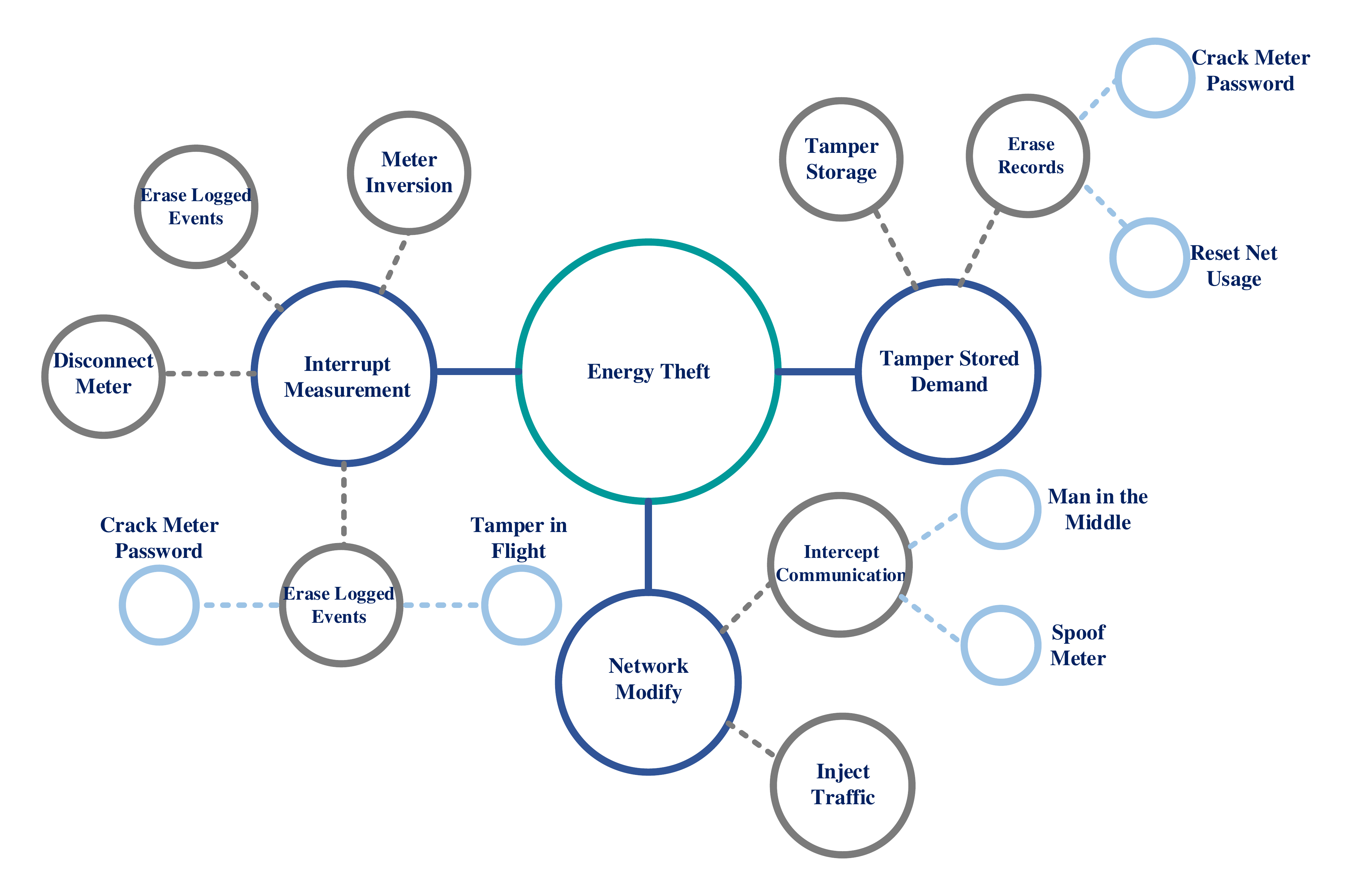}
	\caption{Various Techniques for Energy Theft.}
	\label{Fig:ET}
\end{figure*}

\subsection{Energy Theft Detection}
Energy theft refers to stealing of electricity and also changing of one’s electric power data in order to minimise his or her power bill. It is considered as one of the largest crimes in the United States. There are several ways of electricity theft in a SG. Some of the techniques are erasing logged events, tampered stored demand, tampered meter, disconnection of the meter, etc. as depicted in Fig.~\ref{Fig:ET}.  Previously, power utilities used to send groups to investigate power equipment depending on public reports. The increase in advancement of metering infrastructure made it easier to detect energy theft using data available from smart meters. This gave rise to advanced metering infrastructures (AMI) \cite{gao2019physically}. However, there are many disadvantages that are introduced by AMIs such as being able to manipulate meter readings. This led to the contribution towards feature engineering framework especially to detect theft in smart electricity grids. 

Razavi \emph{et al.} \cite{razavi2019practical} proposed a structure that combines a genetic programming algorithm and finite mixture model clustering for customer segmentation. This aimed to generate feature set which conveys the significance of demand over time. Also comparison among similar households made it reliable in detecting abnormal behaviour and fraud. A wide range of ML algorithms were used. The outcome is outstanding as this method is computationally very practical. With reference to classification algorithm, the Gradient Boosting Machines exceeded all other ML classification models that are used before. This has vital practical signs for electric utilities and can be surveyed better in further research.

Non-technical loss (NTL) in SGs is defined as energy that is distributed but if it is not billed or paid because of energy theft. This has become a major issue in the power supply industry worldwide. In order to detect bypassing of the meter and NTL of meter manipulating at the same time, Kim \emph{et al.} \cite{kim2019detection} proposed power distribution network based model, called as, the intermediate monitor meter (IMM). This model divides the network into finest and independent networks to examine the power flow properly and effectively detect the NTL. An NTL detection algorithm is proposed in order to solve linear system of equations (LSE) that is developed by examining the balance of energy with IMMs and the collector. The authors also stated the hardware architecture of IMMs. This architecture was efficient in terms of time and was able to withstand an accurate detection of at least 95\% accuracy. It was also verified that it finds the morality of consumers and the loss of energy from bypassing that was difficult through traditional detection methodologies

To overcome the loopholes of existing ML based detection methods, Chen \emph{et al.} \cite{chen2020electricity} proposed a new detection method called electricity theft detection using deep bidirectional RNN (ETD-DBRNN), which is used in capturing the internal characteristics and the external association by examining the energy consumption records. Experimenting on the real world data  indicated the proof of this method. Compared with already existing methods, it was found that this method captures the information of the electricity usage records and the internal features between the normal and abnormal electricity usage patterns. The fact is that the value of electricity theft loss should be more associated to the meter readings of energy thieves than to those of moral consumers. Influenced by this fact, Biswas \emph{et al.} \cite{biswas2019electricity} methodically created the problem of electricity theft identifying as a time-series correlation analysis problem. They defined two coefficients to test the doubtful level of the individual consumer’s reported energy usage pattern. Experiments showed that this method improved the pinpointing accuracy when compared to other recent existing methods.

The IoT and AI are two important base technologies making it possible for smart cities. Yao \emph{et al.} \cite{yao2019energy} proposed an energy theft detection scheme using energy privacy preservation in the SG. They used CNNs to detect anomalous characteristics of the metering data from examining a long-period pattern. They also employed paillier algorithm as a security protection for energy privacy. This method showed that data privacy and authentication are achieved together. These results from experimentation demonstrated that the altered CNN model was successful in detecting abnormal behaviours at a high accuracy which up to 92.67\% \cite{yao2019energy}. 

Liu \emph{et al.} \cite{liu2020hidden} presented a hidden electricity theft (HET) attack by utilizing the emerging multiple pricing (MP) scheme. They proposed an optimization problem which aims at increasing the attack profits while avoiding current detection methods to construct the HET attack. They also designed two algorithms to for attacking the smart meters. The authors disclosed and utilized many new vulnerabilities of smart meters in order to illustrate the possibility of HET attacks. The authors proposed several defense and detection measures such as limiting the attack cycle, selective protection on smart meters, and updating the billing mechanism to protect SGs against HET attacks. The proposed countermeasures were able to decrease the attack’s impact at a low cost.

The authors in \cite{buzau2020hybrid} proposed a new thorough solution to self-learn the characteristics and behaviour to detect abnormalities and cheating in smart meters with the help of a hybrid DNN. This model is a combination of a multi-layer perceptrons network and a LSTM network. The results showed that the hybrid neural network performs effectively when compared with the most recent classifiers as well as old DL architectures used in NTL detection. Real smart meter data of Endesa, which is the largest electricity utility in Spain, is used for training and testing of the proposed model. The AMI networks are prone to cyber-attacks where fraud consumers report incorrect electricity consumption which is low in order to reduce their electricity bills in an unauthorized way. To overcome these challenges, the authors in \cite{nabil2019ppetd} proposed a privacy-preserving electricity theft detection (PPETD) methodology for the AMI network. Experimentation on real time datasets was done to evaluate the performance and the security of the PPETD. The results proved the superiority of the proposed model when compared to the existing state-of-the-art to detect fake consumers with privacy preservation, acceptable computation and communication overhead.

The authors in \cite{igrahve2019predicting} modelled energy theft making use of a soft computing approach: swarm algorithm and fuzzy cognitive map (FCM). To design cognitive maps for energy theft parameters, fuzzy logic was used  and to illustrate the weights and concepts’ values, swarm algorithm was used. After being tested using experts’ judgements, it was noted that this model performed satisfactorily better when compared with evolutionary-based FCM. They presented a cost-effective and successful remote detection and identification architecture in order to detect unauthorized consumption of energy. This method detects the fraud consumers in real time by extensive analysis of a large amount of data collected.
 Large-scale simulations using Simulink were performed in order to prove the superiority of the proposed method. Also, three microcontroller-based meters and Simulink environment were used to simulate hardware-in-the-loop simulation. All these experiments proved that the proposed method can efficiently detect false users\cite{halabi2019remote}. 
 
 In the total electricity production of India, more than one fifth of it is lost because of energy theft. Razavi \emph{et al.} \cite{razavi2019socio} made use of district-level data from Uttar Pradesh state of India, over a  period of seven years (2006–2012), and examined the socio-economic predictors of electrical energy theft patterns. Theye deployed a series of advanced ML regression models and explored the temporal-spatial correlation of electricity theft across all the respective districts. The results illustrate that making use of a random forest regression model, 87\% of the variability of loss rate was explained by the existing socio-economic attributes considered in this particular research. They also proved the strength of temporal-spatial correlations of electricity theft across some districts while the average correlation was 0.39 to closest districts and just 0.14 to remote districts.

 The exploitation of information and communication technology (ICT) in electricity grid infrastructures enabled growth in energy transmission, generation, and distribution. Because of costs involved, utility suppliers are only embedding ICT in parts of the grid which gave rise to partial SG infrastructures. Jindhal \emph{et al.} \cite{jindhal2020tackling} argued that making use of the partial SG data deployments can still be effective in solving various challenges such as energy theft detection. They focused on many data-driven techniques to identify energy theft in power grid networks. This technology employed can indicate various forms of energy theft. They also presented case studies to validate successful working of the approach. 
 
 Many challenges, such as electricity theft, were introduced again due to digitalization of power meters which required modern detection schemes and architecture using data analysis, ML and forecasting. Hock \emph{et al.} \cite{hock2020using} demonstrated a multidimensional novel detection approach and architecture to detect tampering of the electricity meters at an early stage by comparing a set of various energy demand time series. This method complemented and enhanced recent monitoring systems which are generally capable of examining a single time series. They aimed to identify electricity theft and presented three data pre-processing methods. 
 This method showed that the metric is vigorous against manipulated data sources. Having more than 90\% detection rates, the authors showed the main reason for using multiple data sources simultaneously while when used individually, it provided smaller value in anomaly detection. This method also showed that different households can be used as data sources to compare avoiding grouping the households based on similarities first.
 \begin{table*}[ht!]
\centering
\caption{Summary on Energy Theft Detection. }
\label{tab3.3}
\resizebox{\textwidth}{!}{%
\begin{tabular}{|l|p{5.2 cm}|p{3.25 cm}|p{7.5 cm}|}
\hline
Ref. &
  Contribution &
  Approaches &
  Key features \\ \hline
\cite{razavi2019practical} &
  Proposed a structure that combines both finite mixture model clustering for customer segmentation and a genetic programming algorithm &
  Gradient Boosting Machines &
  Aimed to generate feature set which conveys the significance of demand over time. Also comparison among similar households made it reliable in detecting abnormal behaviour and fraud \\ \hline
\cite{kim2019detection} &
  NTL detection algorithm is used to solve LSE that is developed by examining the balance of energy with IMMs and the collector &
  NTL detection algorithm &
  IMM-based power distribution network model is used with the concept of unit networks dividing the network into finest and independent networks to examine the power flow properly and effectively detect the NTL \\ \hline
\cite{chen2020electricity} &
  A new detection method called electricity theft detection using ETD-DBRNN, which is used in capturing the internal characteristics and the external association by examining the energy consumption records &
  Deep bidirectional RNN &
  Captures the information of the electricity usage records and the internal features between the normal and abnormal electricity usage patterns \\ \hline
\cite{biswas2019electricity} &
  Electricity theft identifying as a time-series correlation analysis problem &
  Time-series correlation analysis &
  Defined two coefficients to test the doubtful level of the individual consumer’s reported energy usage pattern \\ \hline
\cite{yao2019energy} &
  Proposed an energy theft detection scheme using energy privacy preservation in the SG &
  CNN, Paillier algorithm &
  CNNs to detect anomalous characteristics of the metering data from examining a long-period pattern, also employed paillier algorithm as a security protection for energy privacy \\ \hline
\cite{liu2020hidden} &
  Proposed an optimization problem which aims at increasing the attack profits while avoiding current detection methods &
  Emerging MP scheme &
  Introduced several defense and detection measures, including selective protection on smart meters, limiting the attack cycle, and updating the billing mechanism which aims to decrease the attack’s impact at a low cost \\ \hline
\cite{buzau2020hybrid} &
  Proposed a new thorough solution to self-learn the characteristics and behaviour for detecting abnormalities and cheating in smart meters with the help of a hybrid DNN &
  Hybrid DNN, long short-term memory network and a multi-layer perceptrons network &
  Long short-term memory network and a multi-layer perceptrons network performs effectively when compared with the most recent classifiers as well as old DL architectures used in NTL detection \\ \hline
\cite{nabil2019ppetd} &
  Proposed an PPETD scheme for the AMI network &
  PPETD the AMI network &
  Experimentation on real time datasets was done to evaluate the security and the performance of the PPETD which proves the accuracy of the model in detecting fake consumers with privacy preservation and acceptable communication and computation overhead \\ \hline
\cite{igrahve2019predicting} &
  Modelled energy theft making use of a soft computing approach FCM and swarm algorithm &
  Fuzzy logic &
  Fuzzy logic was used to design cognitive maps for energy theft parameters and illustration of the weights and concepts’ values was done by using swarm algorithm \\ \hline
\cite{razavi2019socio} &
  Deployed an array of advanced machine-learning regression models &
  Advanced ML regression models &
  Explored the temporal-spatial correlation of electricity theft across all the respective districts, proved the strength of temporal-spatial correlations of electricity theft across some districts while the average correlation was 0.39 to closest districts and just 0.14 to remote districts \\ \hline
\cite{jindhal2020tackling} &
  Focused on many data-driven techniques to identify energy theft in power grid networks &
  Data driven techniques &
  Technology employed can indicate various forms of energy theft, also presented case studies to validate successful working of the approach \\ \hline
\cite{hock2020using} &
  Enhanced recent monitoring systems which are generally capable of examining a single time series, aimed to identify electricity theft &
  Multidimensional novel detection approach &
  Demonstrated a multidimensional novel detection approach and architecture for the early detection of tampered with electricity meters by comparing a set of various energy demand time series \\ \hline
\end{tabular}%
}
\end{table*}

We summarize the existing works on energy theft detection in Table \ref{tab3.3}.


\subsection{Energy Sharing and Trading}

Energy management is an important issue concerned with power systems. Energy trading is an energy management technique used to improve the efficiency of the power systems \cite{han2020smart}. It is changing from centralized to distributed manner. But with energy sharing and trading, comes security and reliability issues. Hence it is also necessary that SG attacks are prevented. 

A framework using both DL and DeepCoin, 
a blockchain-based energy framework is proposed by \cite{ferrag2019deepcoin}. A reliable peer-to-peer energy system based on Byzantine fault tolerance algorithm is incorporated in the blockchain scheme. The framework also uses short signatures and hash functions to exploit the energy access and prevent the smart-grid attacks. The DL scheme is an intrusion detection system which uses RNNs for detecting network attacks in the blockchain based energy network. The performance of the framework is evaluated using three datasets and a high throughput is observed demonstrating that the scheme is efficient.

An iterative algorithm for local energy trading between distribution network’s players is presented by \cite{gazafroudi2019iterative}. Results of the simulations based on flexible behaviour of the end-users depicted that shiftable end-users have more dynamic flexibility than that of self-consumption end-users. Also, higher profits for the Distribution Company and the aggregators are achieved by shiftable end-users. It is also seen that the resultant distribution network becomes a sustainable energy system.

A reinforcement learning (RL)-based  microgrid (MG) energy trading scheme is proposed by \cite{lu2019reinforcement}. Based on the  predicted future renewable energy generation, the estimated future power demand, and the MG battery level, the trading policy can be chosen accordingly. The paper also provides a performance bound on the MG utility. Simulation results based on realistic renewable energy generation and power demand data show that this scheme performs better than the benchmark scheme. In a smartgrid consisting of three MGs, it is observed that the proposed scheme increases the MG utility by 22.3\% compared to the benchmark scheme.



\section{DL Use Cases for Demand Response and Smart Grids}
\label{Sec:UseCases}
\begin{figure*}[h!]
	\centering
	\includegraphics[width=0.9855\linewidth]{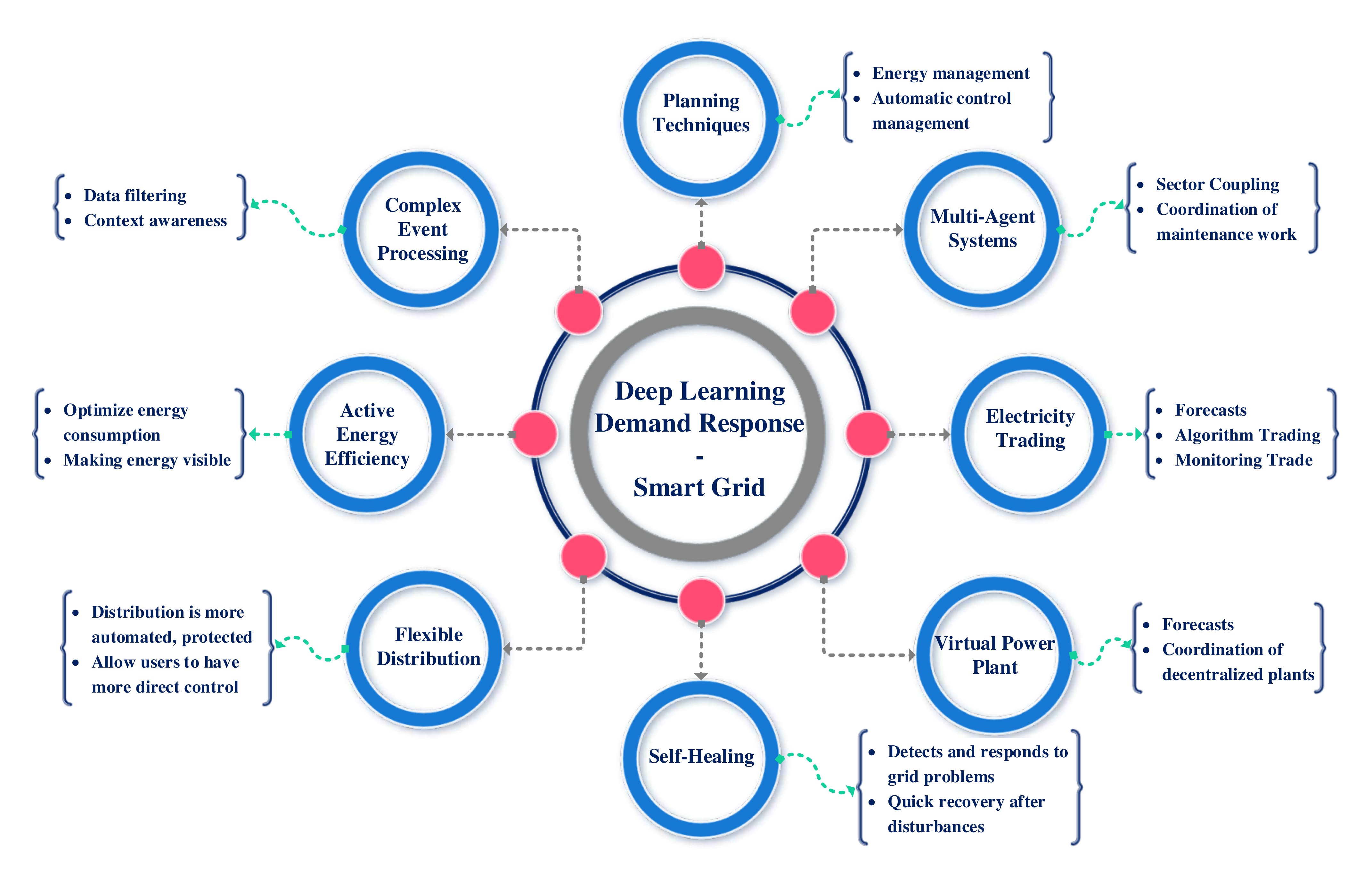}
	\caption{Role of Deep Learning for Demand Response in Smart Grids.}
	\label{Fig:Dr}
\end{figure*}

\begin{figure*}[h!]
	\centering
	\includegraphics[width=0.925\linewidth]{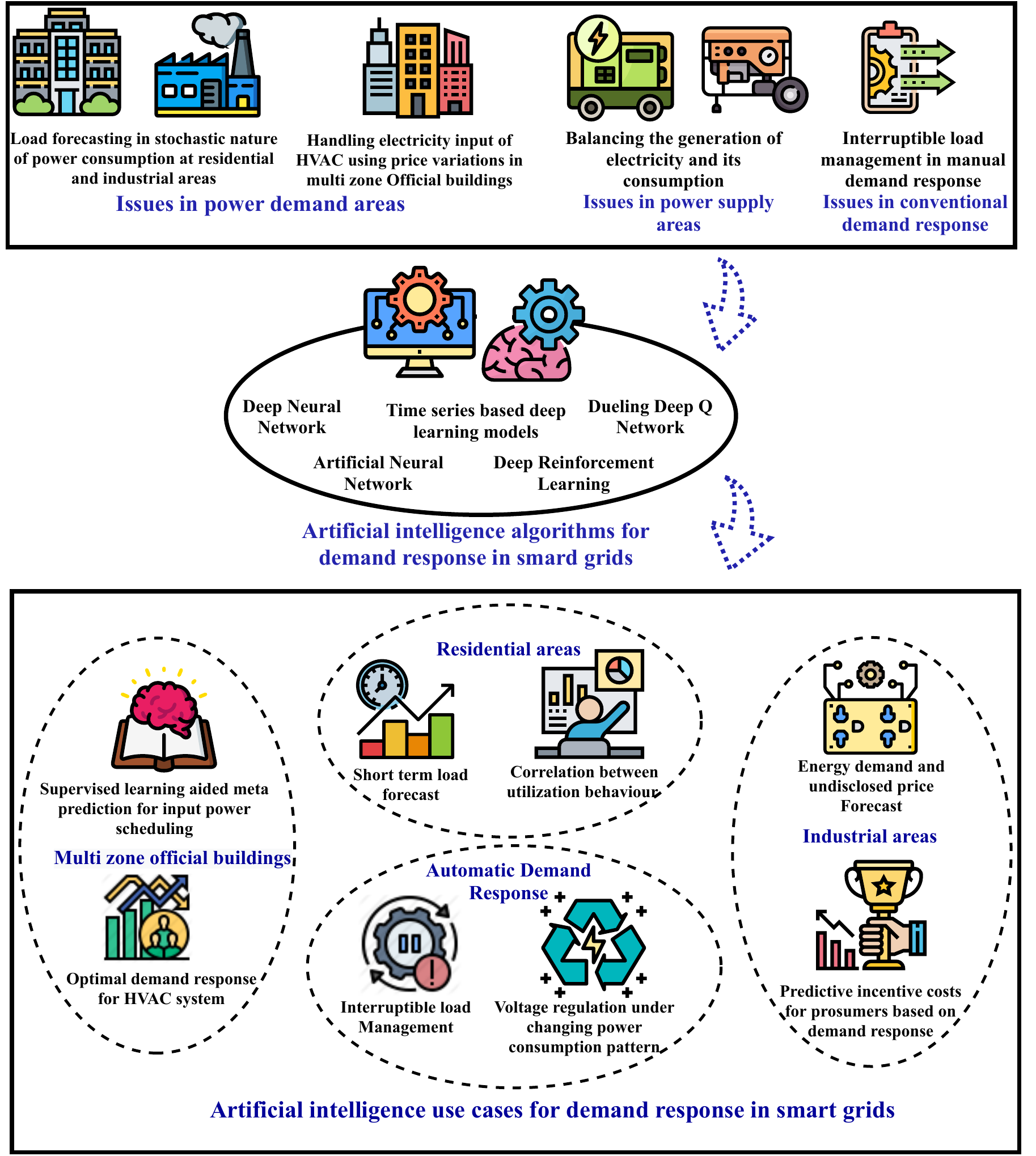}
	\caption{Demand Response Issues in Smart Grids and Usecases of AI Models for the Issues}
	\label{Fig:D1}
\end{figure*}

The use of DL for DR and SGs has attracted much attention from industry and the research community with projects and use case trials. The role of DL models for demand response in smart grids are depicted in Fig. \ref{Fig:Dr}. This include planning, handling multi-agent systems, electricity trading, managing virtual power plant, faster recovery (self healing), optimized energy consumption, flexible distribution and complex event processing \cite{gungor2011smart}.  In this section, we highlight the most popular use cases on the adoption of DL for DR and SGs. Fig. \ref{Fig:D1} depicts the issues concerned with demand response and usecases of AI models to handle them.

\subsection{Incentive based Real-Time DR Algorithm for Smart Grids}
The necessary activity required for soothing the power grids are balancing the generation of electricity and its consumption. There will be an increase in the cost for the producer and consumer if there is any discrepancy in between supply and demand. Several models have been developed so far to balance the energy variation and enhance the grid authenticity.

The first project is an incentive based real time DR algorithm proposed for SG energy systems using DNN algorithm and reinforcement learning. The aim of this algorithm is to provide assistance to the service provider to buy energy resources from its registered customers in order to stabilize the energy fluctuations and improve the authenticity of the grid \cite{lu2019incentive}. A DNN algorithm is used to forecast the undisclosed prices and the demand for energy. Reinforcement learning is applied to acquire the best incentive cost for various customers by examining the net profit of both customers and service providers. 
The service providers are able to acquire the cost from the electricity market and demand for energy from various customers only in the need of the hour because of the intrinsic behaviour of the hierarchical electricity market. To solve these kinds of future unpredictability, DNN is applied to forecast the undisclosed price and demand for energy. The DNN collects new rate and energy demands and predicts the future rate and demand for energy. This activity is repeated many times till the end of the particular day. Further, along with the predicted future rate and energy demands, reinforcement learning is applied to acquire the best incentive rates for various customers by analysing the profits of service providers and customers. The advantages of application of reinforcement learning have several benefits such as model-free, adaptive nature and preciseness. The service provider finds the best incentive price by learning from customers directly and it need not require any prior expertise or predetermined rule on selection of incentive prices. The service providers can obtain the incentive prices independently in an online method remodelled to various customers. Further it considers the flexibility's and uncertainties of the energy system. The entire calculation part of the reinforcement algorithm is dependent on a mapping table and it is very easy to implement this in several real world applications.
The results prove that the proposed algorithm instigates the participation from the demand side, improves the profit of service providers and customers and enhances the reliability of the system by stabilizing the energy resources. In future, the incentive based algorithm can be expanded to extensive capacity resource market framework including the Grid operators and several service providers.

\subsection{Load Forecasting Framework for Residential Areas}
The next DL project related to SG is short-term load forecasting for residential areas. In order to improve the efficiency of the electrical power system, DR in residential area is essential. The basic task in DR system is forecasting the short-term load accurately. Several research work has been done on short term forecasting of accumulated load data. But forecasting load for single residential users is challenging because of its stochastic and dynamic nature of power consumption behaviour of individual users. To address this issue, a framework for STLF for residential area was proposed using spatio-temporal correlation in load data and by the application of DL algorithms \cite{hong2020deep}. STLF helps to forecast the demand of the users in the forthcoming future and also provides recommendations on DR in residential area. It also fulfils the electricity demand of the users and minimizes the danger of outages. The spatio-temporal correlation applied in this framework defines the spatio correlation amidst the power consumption practices of various kinds of devices and the temporal correlation among the archival power consumption practices and future power consumption practices.
In this framework, the spatio temporal correlation between various kinds of power consumption practices were explored to enhance the SLTF's performance. Several time series analyses were conducted in order to define the power consumption practices of various applications and also their spatio temporal interrelations. A method using DL algorithm with iterative ResBlocks for STLF was proposed to find the correlation between the utilization behaviours. The iterative ResBlocks based method was applied to acquire knowledge in deep and shallow features of input data. The proposed framework constitutes four steps which includes data collection, preprocessing, training and load forecasting. The data collection step gathers units from smart meters which provides information about power consumption of various devices for each customer. Data preprocessing step performs data cleaning, integration and transformation in order to enhance the data quality of STLF framework. Training step applies a DL algorithm with an iterative ResBlock method to find the spatio-temporal correlation between various power consumption practices. Finally after performing preprocessing and training steps, the proposed model computes the forecast values of individual users.

\subsection{Optimal Demand Response in Multi-zone Official Building}
Another interesting DL use case on DR is finding optimal DR of heating, ventilating, and air-conditioning (HVAC) system in multi-zone official buildings.  HVAC loads constitute nearly 30\% of the power consumed by an official building. Several steps have been taken by distribution system operators for DR projects to manage the electricity input of HVAC systems indirectly and directly using electricity rate variations and communication links respectively. Many studies have been done on DR for HVAC systems. The developed models consist of several factors to be gathered from parameter estimation methods and it is a time consuming process. To overcome the problem of time consuming models and parameter tuning, ML algorithms have been used in DR for HVAC systems. 
A new approach for optimal DR of HVAC systems in a multi zone commercial building was proposed using supervised learning methods \cite{kim2020supervised}. The data obtained from normal building operating circumstances were applied to train ANN. The ANN with time delayed input data, feedback loops, various activation functions and several hidden layers was applied to find the variation in temperature in each zone when there is change in the power supply of the HVAC system. In the supervised learning algorithms, the ANNs are modelled and trained using the past data gathered. The ANN methods are iterated using linear equations and combined to form an optimization problem for rate based DR systems. The issue is resolved for several building thermal situations and electricity rates. The obtained solutions are applied to train a DNN model in order to find the optimal DR method namely supervised learning aided meta prediction (SLAMP). The SLAMP method schedules the input power of the HVAC system directly for building thermal conditions and electricity rates for the next 24 hours. This helps to reduce the time for computation and provides a curve for rate and optimal demand helping the distribution system operator to optimize the load for HVAC through power pricing.

\subsection{Demand Response Management of Interruptible Load using DRL}
Another promising DL project is DR management of IL. Conventional DR systems need manpower to operate the machinery or fix the operating parameters manually. This does not guarantee the response pace of the users and reliability of implementation of DR. Automatic DR is one of the powerful technologies of SG and it implements the DR system with the help of automatic systems for scheduling the load and to increase the efficiency of energy.
Several DR models have been developed and are categorized into incentive based and price based models. Price based DR models refers to the user behaviour with respect to the modification in electricity consumption rates. Incentive based DR models represent the behaviour in order to acquire profitable incentives. The important task of incentive DR is interruptible user load in peak load situations or in case of emergencies to achieve a fast response and enhanced demand side strength. Model based optimization methods associated with IL needs physical and mathematical model explicitly and it leads to difficulty in adapting to practical operating circumstances. Therefore a model free method based on DRL was designed with a DDQN in order to optimize the management of DR of IL for time of use tariff and changing power utilization patterns \cite{wang2020deep}. 
An architecture for DDQN based automatic DR was constructed initially to provide a solution for practical applications of DR. The problem of management of DR in IL is described as a Markov decision process to acquire utmost long term profit. The DDQN based DRL algorithm helps to solve the problem of Markov decision process with utmost progressive reward. The actual grid state is directly mapped to the DR management policy in the proposed DDQN based DRL algorithm and attains the aim of regulation of voltage and minimization of overall operation rate of distribution system operators under changing power consumption patterns.

\section{Challenges, Open Issues, and Future Directions}
\label{Sec:Challenges}
The SG technology has made its remarkable effort in alleviating the major problems faced by the traditional electric grids and the conventional way of electricity distribution. SG technology utilizes the ML techniques and more specifically, its subset, the DL approaches for handling high-dimensional data and ensuring effectiveness in data transactions throughout the energy supply chain. Also, SG technology focuses mainly on consumer satisfaction, thereby making the consumer a prosumer (producer and consumer) by effectively managing the power consumption of consumers and enabling energy sharing facilities. SG technology accomplishes this by allowing the bidirectional communication between the consumer and the utility instead of uni-directional communication as in traditional power grids. Though SG technology solved the major problems in conventional electricity, it has many challenges in handling varied stakeholders, data synchronization in the energy supply chain, regulatory requirements, and so. Furthermore, it has some open issues in load DR management and future challenges for the betterment of future electricity needs. This section presents the various research challenges, available issues and future directions in SG technology by utilizing DL approaches.

\subsection{Challenges}
The various challenges in a SG namely uncertainty in different types of energy sources (like solar and wind), security and stability of large scale power systems, load forecasting during peak time, decision control problems, DR based load decomposition and balancing (during peak and off-peak period), equipment health monitoring systems, energy management in microgrids and microgrid coordination, EVs management and malicious data injection require mandate attention for effective utilization of SG technology. Some of the challenges and feasible technical solutions are presented here. 

\subsubsection{Dynamic pricing for demand response in microgrids} 
A microgrid is a local electricity source connected to the SG environment, which can operate autonomously. Micro-grid adopts the dynamic pricing technique where the service provider will act as an intermediary (brokerage) between the consumer and the utility company by selling the electricity purchased from the utility company to the consumer. Though the service provider uses dynamic pricing to manage microgrids, the uncertainties in microgrids and inaccurate customer details (consumer load demand level, customer purchase pattern and usage patterns),  makes it more challenging to determine the pricing based on consumer’s future behaviour. On the other hand, consumers face the problem of energy consumption scheduling incurred through uncertainties in electricity price and their usage. RL based multi-agent model was proposed in \cite{kim2015dynamic} to learn both the consumer’s energy consumption and the service provider’s dynamic pricing based on future behaviour. Also, they have used the post-decision state learning algorithm to enhance the consumer’s learning rate of minimizing their energy cost.
Load flexibility analysis on the demand side: The most challenging task in DR is to predict the energy flexibility on the consumer side. Non-intrusive load monitoring is used to determine the energy consumption of the individual appliances at the consumer end and helps to analyze the consumers’ energy consumption behaviour in real-time \cite{zhang2018review}. The major problem with this appliance/device energy consumption analysis is randomness and partially observed results. DL methods are used in the literature for load prediction to assist DR. RNN is used for consumer classification based on their energy consumption pattern \cite{tornai2017recurrent}. Device-based RL technique was proposed in \cite{wen2015optimal} to determine the energy consumption of device clusters. Based on the energy consumption profile obtained from the device level, DR problem of energy consumption scheduling at the aggregated level is resolved. Q-learning algorithms are used for online scheduling of resources in residential DR \cite{mocanu2018line}, and CNN is used to determine the DR problem with partial observability \cite{ruelens2016residential}.

\subsubsection{Load forecasting in smart grids}
The energy flexibility analysis can be done effectively through accurate prediction of future load, i.e., the future energy demand of consumers based on their previous usage. This will ensure sustainable energy systems with minimal waste and fair pricing. This energy prediction depends on various factors like climatic changes and transition probability. Furthermore, a minute load forecasting error may cause huge economic loss. Therefore an effective load forecasting in SG is challenging and should be addressed efficiently. DNN algorithm (namely long short term memory architecture) and ANN-based time series models can be used for accurate prediction of energy in residential and commercial buildings \cite{bagnasco2015electrical}. Stochastic models for building energy prediction developed in \cite{mocanu2016deep} outperforms the existing ANN and DNN models. Though the stochastic models provide an accurate prediction of aggregated power consumption, the overall performance can be improved by tuning the learning rate. Bayesian DL based model was developed in \cite{yang2019bayesian} to predict the load by quantifying uncertainties among different customer groups. The model also uses a clustering-based pooling method to avoid overfitting and improving prediction performance. Furthermore, an aggregated model obtained by integrating forecasting results of load dispatch model and power load and PV output model in community microgrid was developed in \cite{wen2019optimal} to stimulate supply-demand balance. PSO algorithm is used for the optimization of load dispatch between the grid and the connected community grid. The aggregated model has shown more remarkable improvement in reliability and costs. Therefore, aggregation of DL models with optimization techniques provide better forecasting results.

\subsubsection{Equipment health monitoring}
SG technology sources its information from diversified pieces of equipment (inclusive of devices used for power consumption meters and power generators). Monitoring the status of the equipment in a large scale power system is a vital need for the effective functioning of the system and easier maintenance. The faults detected earlier will avoid unwanted chaos in electricity distribution and ensure satisfied consumers. Consequently, this has a greater impact on retail electricity pricing. Technical fault detection in wind turbines demanding impulsive maintenance is essential. DL approaches were analyzed in \cite{helbing2018deep} for wind turbine condition monitoring by detecting the faults at earlier stages. From the literature, they suggest that unsupervised learning algorithms are more prevalent for fault detection than supervised algorithms.

\subsubsection{Cyber attacks in bidirectional energy trading}
SG technology enables bi-directional communication between the consumer and the utility company thus focusing on prosumer instead of the consumer.  The bi-directional communication handles different types of sensitive data between the two entities. A SG uses information and communication technology for most of its activities which makes it vulnerable to various cyber-attacks. Also, the intruder to steal the energy or to deceive the consumer energy profile can induce multiple cyber attacks. False data injection is one of the sombre threat to SCADA which an embedded system that allows the business organizations to control the industrial processes locally as well as remotely with GUI (enabling human to interact directly with various devices). A DL approach namely, conditional DBN was employed in \cite{he2017real} to detect the false data injection attacks in real-time based on the historical data and live features captured for detecting the electricity theft. A DL based defence mechanism for cyber attacks was proposed in \cite{wang2018deep}. The cyber attack model is designed based on a different state of the system as the attack patterns vary depending on the energy consumed to deceive the information. The model can effectively detect the anomaly states of the system. Deepcoin, an RNN based intrusion detection system to detect fraudulent transactions and network attacks blockchain-based energy trading\cite{ferrag2019deepcoin} using hash functions and short signatures. Deepcoin avoids the double-spending attack, denial of service attacks, false data injection and brute force attack. DL based cyber-physical system for identifying and mitigating false data injection attack in SGs was proposed in \cite{wei2016deep}. Wide and deep CNN was designed in \cite{zheng2017wide} to detect electricity theft based on abnormal energy consumption pattern of malicious consumers. The centralized SDN based SGs was in \cite{parra2019implementation} developed for deep packet inspection. The DL models are used for classifying the cyber attacks by their family using deep packet injection. The models can successfully classify insider attacks, ransomware attacks and denial of service attacks.
\subsubsection{Energy management with EVs: EV scheduling}
Electric motor services are emerging, and these services need greater attention in the automobile industry. EV is a distributed energy source that can be used as an alternative for the internal combustion engine for electricity transportation. EVs are used in a vehicle to grid (V2G) and vehicle to home(V2H) communications of smart grid technology. The integration of EVs with the electric grid will increase its power quality, efficiency and performance specifically during peak hours. Some of the services offered by EVs are peak power shaving (balancing load during demand), spinning reserve (increasing power generating capacity), power grid regulations and load levelling \cite{ tan2016integration}. Recharging the EV batteries impose a greater challenge in maintaining EVs (because of more number of uncertainties). Sometimes charging of EVs will be very frequent and may consume more time. Scheduling the charging time for EVs should be highly focused on providing a balanced load in all the EVs without collision (simultaneous charging of more EVs at one particular time may exhaust the power system). DNN was employed in \cite{8299470} to make real-time decisions on charging EVs (during the connection) to reduce the vehicle power cost based on the historical data on the connections. A decision is based on various parameters such as demand series, environment and pricing. Furthermore, the EV charging load was predicted based on the traffic flow forecast in \cite{9055130} using CNN and scheduled using the queuing model. The arrival rates of EVs are calculated based on the traffic flow intervals and historical data. ML algorithms were employed in \cite{ vanitha2020machine}, select a charging station with optimal waiting time and faster-charging ports at charging stations. Of various DL techniques for EV load forecasting, LSTM provides better results with less forecasting error \cite{zhu2019electric}
Therefore, while designing a prediction system for the smart grid, the challenges, as mentioned above, should be considered to provide the sustainability and attack free energy trading.

\begin{table*}[t]
\centering
\caption{Research Challenges and solutions.}
\label{tab1}
\begin{tabular}{|p{0.5cm}|p{2.1cm}|p{4.5cm}|p{4.5cm}|p{3.5cm}|}
\hline
\textbf{Ref.} &
  \textbf{Challenge} &
  \textbf{Application} &
  \textbf{Solution} &
  \textbf{Benefits} \\ \hline
\cite{kim2015dynamic} &
  Dynamic demand pricing and consumer energy consumption scheduling &
  RL based multiagent learning   algorithm in microgrids &
  Both the consumer and service   provider learns the strategy without apriori information on dynamics of the   microgrid. &
  Efficient energy consumption   scheduling and better pricing capabilities for the service provider. \\ \hline
\cite{ wen2015optimal}, \cite{ tornai2017recurrent}, \cite{mocanu2018line}, \cite{ruelens2016residential} & Consumer energy consumption scheduling
   &
  DL, RL, CNN, Q-learning and RNN in   consumption scheduling at small commercial and residential buildings &
  Determining the energy consumption level of   individual appliances to solve the consumer energy consumption scheduling &
  Accurate consumer energy consumption scheduling \\ \hline
\cite{wang2018deep} &
  \multirow{5}{*}{Cyber attacks in SG} &
  DL based cyber attack model &
  Determines the anamoly states of   the system &
  Safe energy trading \\ \cline{1-1} \cline{3-5} 
\cite{ ferrag2019deepcoin} &
   &
  RNN and blockchain (hash and short   signatures) for energy exchange &
  Fault-tolerant energy transaction   detects the intruders during trading &
  Safe, fault-tolerant,   privacy-preserving energy trading and high throughput \\ \hline
\cite{mocanu2016deep} &
  Load forecast in SGs &
  ANN, CNN, stochastic models for   building energy prediction &
  Energy consumption prediction by   considering various dynamically changing parameters &
  Accurate energy load prediction \\ \hline
\cite{helbing2018deep} &
  Equipment health monitoring &
  DL for wind turbine condition   monitoring &
  Unsupervised learning algorithms   are more prevalent than supervised learning algorithm &
  Effective maintenance \\ \hline
\cite{8299470} &
  \multirow{5}{*}{EV Charging in SG} &
  DNN for charging of EVs &
  Makes real-time charging decisions   based on historical data on connections. &
  Reduced EV charging cost \\ \cline{1-1} \cline{3-5} 
\cite{9055130} &
   &
  CNN and queuing model for   scheduling EV charging &
  The EVs are scheduled for charging   based on the traffic flow interval and historical data. &
  Effective scheduling of EV   charging by learning uncertainities. \\ \hline
\end{tabular}%
\end{table*}

\subsection{Open Issues}

SG technology has started using various DL approaches to sort out its varied challenges. But the application of DL in SG technology is fewer when compared to other complex domains. Some of the open issues are listed in this section. 
\subsubsection{Procuring Labels from Inconsistent Operational Data}The primary reason for the less significant application of DL in the SG technology is complexity in procuring labels with a high number of parameters. This is because of the inconsistencies in operational data.  The various components in the SG, namely master control units in the power plants connected to the remote terminal units will deliver the power energy to the consumers (through intelligent electronic devices for secured energy delivery and monitoring). A smart meter, one of the significant components of the SG environment, acts as an intermediary between the utility grid and the consumer. It has many modules such as timing, communication, metering, indicating, encoding, power and control modules. Tampering of data during data communication among these modules may lead to inconsistencies in power supply, power pricing and loss of sensitive information.

\subsubsection{Security concerns with enormous data accumulations} Through smart grid technology, energy sector utilizes the various IoT devices, wireless network communication and evolving technologies (like the cloud) more data is accumulating. So big data analytics can be preferred for handling the large volume of data at power generation demand-side management, microgrid and renewable energy sources \cite{ zhou2016big}. Although big data serves in managing energy data effectively, big data processing systems are facing a lot of challenges in handling diversified, and complex energy bigdata \cite{tu2017big}. Furthermore, big data has various loopholes in ensuring the security and privacy of more sensitive consumer energy profiles.

\subsubsection{Security Issues with Vulnerable Centralized Controllers} With technological advancements, the SG environment is prone to a variety of new cyber attacks.   Henceforth various security solutions were enforced on SG environments such as SDN for secured data communication. The major issue concerned with SDN based SGs is the central point of failure caused by its vulnerable centralized controller and grid devices \cite{parra2019implementation}.  Also, SDN based deep packet injection solutions cannot account the mutability of the attacks because they are resource-intensive. Therefore, other alternatives should be explored.

\subsubsection{Security issues with enormous data accumulations} Through smart grid technology, energy sector utilizes the various IoT devices, wireless network communication and evolving technologies (like the cloud) more data is accumulating. So big data analytics can be preferred for handling the large volume of data at power generation demand-side management, microgrid and renewable energy sources \cite{zhou2016big}. Although big data serves in managing energy data effectively, big data processing systems are facing a lot of challenges in handling diversified, and complex energy bigdata \cite{tu2017big}. Furthermore, big data has various loopholes in ensuring the security and privacy of more sensitive consumer energy profiles.

\subsubsection{Incomplete Energy Acquisition} Incomplete energy acquisition in microgrids (brokerage between consumer and utility grid) is one of the prominent issues between consumer and utility companies (or service provider).  So, the utility company and the consumer must know the strategy adopted by the service provider without any prior information about the dynamics of microgrids \cite{kim2015dynamic}.

\subsubsection{Inaccuracies in electric load forecast} Accurate electric load forecast is an emerging issue in SGs. Although many different types of computational methods were employed for electric load prediction, several issues like handcrafted features, impotent learning, limited learning capacity and inability to handle non-linear data. So ML algorithms were used to solve these issues, but those algorithms were not fit for larger data and a small error in forecast led to tremendous effects. Moreover, microgrids have a lot of variation in electricity load. Therefore, an efficient load forecasting model with the feature engineering and optimization technique is required for faster and error-free forecasting. A hybrid model integrating DL and heuristic optimization techniques (genetic wind driven optimization) was developed in \cite{hafeez2020electric} with better accuracy than the existing model. But the model was not evaluated for the real-time data. Scheduling of EVs for charging and EVs load forecast must include various parameters (historical data and real-time data) for making wise decisions and optimal solutions. Therefore, this issue can be resolved by integrating one or more DL techniques with optimization algorithms.

Though the evolving DL models for the SG are providing better solutions, yet to design an efficient system for SG forecasts must address the above mentioned issues.

\subsection{Future Directions}
SG technology utilizes information and communication technology for the automation of various processes in the utility grid, prediction of future trends and offering effective services to the consumer. Utilization of rapid advancements may be prone to a variety of issues in all aspects, as mentioned in the above sections. Some of the future aspects of sustainable energy services through the SG is discussed in this section. 

\subsubsection{Blockchain for smart grid technology}
Security issues concerned with energy big data processing systems can be resolved with the help of blockchain. The decentralized blockchain would be an essential candidate for handling the energy big data \cite{9018104}. The heterogeneous data from diversified sources of energy big data can be stored in the blocks, and the blockchain capability of representing any of the data in a unique format will make the data structured and predictable. They were, therefore reducing the complexity of big data processing systems. Also, blockchain through smart contracts provides secured and privacy-preserving data transactions. A lightweight blockchain framework can be deployed for SG with DL, thereby segregating the metadata by maintaining a side chain for enhancing security and privacy in the energy transactions. Besides, the double-spending attacks concerned with blockchain framework can be alleviated by strengthening the blockchain security. Blockchain usage is still limited as the mining process in blockchain consumes more resources. So, the edge computing, which brings computation closer to the data source by offloading the mining activities, can be integrated with blockchain for low-latency response in energy transactions \cite{ gai2019permissioned}. Henceforth, integration of big data analytics (for volume and veracity of data), blockchain (for integrity and privacy-preserving transaction) and edge computing (faster energy transactions) in smart grid technology, will ensure secured and sustainable energy management services.

\subsubsection{Edge AI for SG technology}
Edge AI, the convergence of edge computing and AI is an emerging solution to reduce delays in performing predictions on load forecasting, energy pricing and demand-side response management. Edge AI, intelligence at the edge helps in making real-time perceptions based on the data generated from the edge devices (like smart meter) using AI algorithms processed at the SG edge devices. For instance, the customer usage pattern can be predicted based on the data accumulated at smart meters \cite{samie2019edge}. Furthermore, edge AI can be utilized for effective power resource optimization in a distributed manner, load forecasting, fault identification and diagnosis at appliances level (the edge devices). This, in turn, will be an added advantage for the future grid, which dramatically reduces the peak demand, saves enormous energy with increased revenue and effective customer service. As discussed in challenges, DL models on cloud platform play a vital role in combatting security issues in SG as well as in blockchain-based energy network. Though DL models serve its best, offering varied services at smart grid can be further improved with edge AI. Through the integration of edge AI with DL enabled SG network, huge delays in complex computation and task offloading can be alleviated by moving the computation at edge devices.  Furthermore, distributing intelligence computation at the edge devices with identity authentication will secure and strengthen the SG. The distributed intelligence, together with anomaly detection (via identity authentication before processing) at the edge, will combat the cybersecurity issues. On the other hand, the edge AI will ensure sub-second latency response in energy transactions. 
\subsubsection{Quantum Computing for SG}
Quantum computing (QC) introduces a new quantum processing unit, namely qubit, which allows the quantum computer to explore simultaneously diverse solutions to a given optimization problem before disintegrating to an optimum solution. Also, through quantum entanglement property of QC, the different qubits can sense what happens to other qubits without passing any signalling messages. However, DL models provide various insights for energy optimization from hybrid power systems in the energy industry. These models must consider multiple data from heterogeneous sources to find an economical combination that can combat energy demands with an acceptable level of security.QC is an appropriate technology for smart grid energy optimization from distributed and heterogeneous sources. As the number of IoT devices and energy generators increases the security risks in various smart grid management activities such as attacks during data accumulations from varied sources and energy transactions during bidirectional trading. The quantum cryptography can be employed to secure SG from these attacks. A lightweight quantum encryption scheme was implemented in \cite{li2019lightweight} by integrating the quantum cryptography and one time pad to ensure secure power data transmission in the smart grid.  Furthermore, quantum cryptography can be integrated with blockchain for secured energy trading from multiple sources.

\subsubsection{5G and above technology for faster energy trading}
The low latency and ultra-reliable 5G networks operating at higher frequencies support machine to machine communication will be benefiting the scalable smart grid technology (as it requires gigabytes bandwidth, connecting to innumerable IoT devices and higher reliability). One of the primitive challenges with SG enabled DL is larger resource consumption (network bandwidth) for processing the energy data accumulations and EVs management during energy trading. Upon integration of ultra-reliable, 5G with SG enabled DL can fasten the data processing and can improve the performance of the DL models with better accuracy. The integration of 5G with smart grid technology enhances the availability (DR management), fault tolerance, cybersecurity and self-healing capability of the grid. Furthermore, it assists effective and faster V2G communications in the smart grid environment through EVs \cite{de2019security}. Although 5G offers indisputable benefit to smart grid, it may lead smart grid vulnerable to more cyber attacks, incompatibility of older devices will require replacement, more device dependency and overcrowding risk at frequency spectrum. To overcome these limitations, sixth-generation (6G) network with higher data rate, sub-second latency response and lesser processing delay \cite{khan20206g}. 6G technology includes various key enablers, namely edge intelligence, homomorphic encryption, ubiquitous sensing, cognitive radio, network slicing and blockchain. 6G integrates these key enablers, emerging ML schemes (federated learning and so on), high-end communication and computing technologies to deliver novel smart applications. Seamlessly, smart grid technology can have optimal, secured and improved performance in energy management.

Though the blockchain, edge AI, QC, 5G for future grid discussed here adds different benefits for smart grid technology individually, they still have their pros and cons. This can be resolved by the optimal integration of technologies overcoming their drawbacks and strengthening the future smart grid technology.

\section{Conclusion} 
\label{Sec:Conclusion}
In this paper, we have presented a survey on the state-of-the-art DL solutions for SG and DR systems. Particularly, we have focused on reviewing four important themes such as electric load forecasting, state estimation, energy theft detection, energy sharing and trading.  A number of use case and projects have been presented to illustrate the need of DL solutions in SG and DR systems. We have also discussed a number of challenges and potential directions on the use of DL. It is our hope that this work will be serving as a source reference for researchers and engineers interested in DL and SG systems, and drive more great researches in the future.

\balance


\end{document}